# Constructing Dynamic Master Logic Models as Knowledge Graphs for Complex System Diagnostics Using Retrieval-Augmented Large Language Models

Saman Marandi [a], Yu-Shu Hu [b], Mohammad Modarres [a]

[a] Center for Risk and Reliability, University of Maryland, 0151F Glenn L. Martin Hall, Building 088, College Park, MD 20742, United States
[b] DML Inc., Hsinchu, Taiwan
**Corresponding Author:** smarandi@umd.edu
**E-mail addresses:**
smarandi@umd.edu (S. Marandi), modarres@umd.edu (M. Modarres), huyushu@dml.com.tw (Y.-S. Hu)

## Abstract

Dynamic Master Logic (DML) provides a hierarchical framework for representing system behavior by linking functional objectives to underlying structural elements. Unlike event-based approaches such as Fault Tree Analysis, DML captures system objectives, functional dependencies, and unanticipated failure scenarios within a unified functional hierarchy. However, DML construction typically relies on expert interpretation of technical documentation, limiting scalability for complex systems. This study presents a framework for automated construction of DML models from system descriptions and their representation as Knowledge Graphs (KG-DML), leveraging Retrieval-Augmented Generation and Large Language Models as the enabling tools. Building upon prior work on small-scale systems, the present work introduces automated construction and evaluation of KG-DML models, extending the paradigm to substantially larger and more complex system. The proposed approach performs model construction across the hierarchy using targeted retrieval, preserving hierarchical dependencies and explicit logical relationships. The resulting KG-DML supports diagnostic reasoning, safety assessment, and graph-based analysis through upward failure propagation and downward dependency tracing. The framework is evaluated through a multi-level validation methodology that combines layer-specific precision and recall, logical gate consistency analysis, and an aggregate integrity metric capturing both element-level accuracy and system-level structural completeness. Application to the Low-Pressure Coolant Injection system of a decommissioned Boiling Water Reactor demonstrates consistent reconstruction across repeated runs and highlights factors influencing model stability. The results show that automated KG-DML construction can transform technical documentation into executable functional models for diagnostic and reliability analysis.

**Keywords:** Dynamic Master Logic, Functional Modeling, Knowledge Graphs, Fault Diagnosis, Retrieval-Augmented Generation, Large Language Models

## 1. Introduction

Modern engineered systems are often composed of many interacting components organized across multiple levels of abstraction. Therefore, modeling these systems accurately requires representations that can account for complex interactions and dependencies. Traditional diagnostic models such as Event Tree Analysis (ETA) [1], [2] and Fault Tree Analysis (FTA) [3], [4] rely on event-driven representations in which faults or initiating events are traced through predefined causal sequences. While effective for known failures, event-driven models require exhaustive fault-event listings, making them inherently incomplete when failure sequences are not anticipated during model development. As systems grow in complexity, tracing failure pathways also becomes increasingly infeasible using event-based logic alone.

Functional modeling offers an alternative approach for complex systems. Functional models describe what the system is intended to accomplish, how objectives are decomposed into functions, and how those functions are realized through interactions among components[1] [5]. By focusing on functions and their dependencies rather than individual events, functional models provide a clearer representation of how component behavior contributes to system-level objectives under both normal and abnormal conditions [6].

Dynamic Master Logic (DML), introduced by Hu and Modarres [7], is a functional modeling framework that represents system behavior through a hierarchy linking functional objectives to underlying structures. DML has been applied in reliability and safety analysis to support reasoning about system behavior and failure mechanisms [8], [9], [10]. In practice, however, constructing DML models has largely relied on manual, expert-driven processes, where developers analyze system descriptions and specifications to interpret system behavior and construct the model. This limits scalability and restricts the use of explicit functional representations needed for reliable reasoning over complex engineering systems documented through extensive technical documentation.

Recent advances in Artificial Intelligence (AI), particularly Large Language Models (LLMs), create new opportunities to support the construction of functional models. Large language models are effective in tasks such as semantic interpretation, text summarization, and generation [11], making them promising tools to help engineers construct functional models from technical documentation. However, LLMs also exhibit important limitations, including inconsistent reasoning, lack of persistent structure, and the potential to generate unsupported information [11], [12]. Recent research has further highlighted the challenge of enabling language models to reason explicitly about actions and their consequences [13]. This limitation is particularly significant in engineering diagnostics that rely on functional models, where accurate reasoning requires not just recognizing relevant information but understanding the functional relationships that govern system behavior. Without an explicit structured representation, an LLM cannot reliably reason about which functions support which objectives or how a component failure cascades through the system. Dynamic Master Logic addresses this limitation by linking objectives, functions, and dependencies into an explicit structured representation that supports traversal and enables LLM agents to reason about functional dependencies and the consequences of component failures, rather than relying solely on generalized language model knowledge. Consequently, accurate DML models provide an explicit knowledge representation that enables grounded LLM-assisted diagnostic reasoning over complex engineering systems.

In our prior work [14], we explored the use of LLMs to construct a Knowledge Graph (KG)-based DML representation for a small-scale system. That study demonstrated the feasibility of generating explicit functional representations capable of supporting LLM-assisted diagnostic reasoning, but did not address scalable construction from large and document-intensive system descriptions. In addition, the evaluation framework was relatively simple and did not systematically assess hierarchical consistency or structural completeness. As systems grow in complexity, these limitations motivate the need for a scalable and rigorously evaluated approach. More importantly, reliable diagnostic reasoning depends on accurate and complete KG-DML models that explicitly capture functional objectives, dependencies, and failure propagation. In practice, constructing such models has relied on manual expert effort, limiting their scalability.

To enable scalable construction of explicit engineering functional representations for grounded diagnostic reasoning, this study presents a framework for automated construction of DML functional models from engineering documentation. Retrieval-Augmented Generation (RAG) is employed as the underlying extraction mechanism to support scalable retrieval from large system descriptions. The framework constructs KG-based DML models that closely reflect expert-derived DML structures using information retrieved from authoritative engineering documentation, though the resulting models should be regarded as LLM-informed and require validation against domain expertise. The model is constructed sequentially along the DML hierarchy, with each layer generated from documentation passages retrieved based on the semantic definition of the layer and the parent nodes established in earlier stages. The resulting model is represented as a KG [15] referred to as a KG-DML, which preserves the hierarchical and logical structure of DML.

[1] In complex systems, components refer its active and passive constituents of the system, they could be complex such a subsystem composed of multiple components in a train of the system, or a simple component.

Since diagnostic reasoning in the proposed framework is performed through deterministic graph traversal and Boolean propagation over the KG-DML, the structural quality of the KG-DML directly determines the quality of the resulting diagnostic inferences. During interaction, the LLM agent is responsible only for interpreting user queries and invoking the appropriate graph-based tools. Errors such as missing elements, incorrect relationships, or incomplete dependency paths can therefore propagate through the model, leading to inaccurate or misleading diagnostic conclusions. Therefore, the proposed evaluation framework examines both layer-specific accuracy and overall structural consistency, combining these into an integrity score that reflects the quality and completeness of the constructed KG-DML.

From a reliability engineering perspective, the proposed framework represents a step in the evolution from manually constructed DML models toward automated, documentation-driven construction, extending the KG-DML paradigm while preserving the functional hierarchy and logical dependency structure that distinguish DML from conventional KG approaches. The resulting executable KG-DML supports diagnostic reasoning, safety assessment, and resilience-oriented analysis through explicit representation of functional dependencies. This work makes three main contributions. First, it introduces an automated approach for constructing DML models from system documentation, leveraging retrieval-augmented LLMs as enabling tools, enabling functional decomposition and dependency identification from unstructured text. Second, it extends our previous small-scale approach to support construction for substantially larger, document-intensive systems. Third, it develops a structured, multi-level evaluation framework for assessing the constructed KG-DML, combining layer-level accuracy metrics with an integrity measure of structural completeness and consistency. The novelty of this work lies not in the individual use of established technologies such as RAG, LLMs, schema-constrained extraction, or graph databases, but in their integration into a coherent framework for automated DML construction from engineering documentation, together with an executable KG-DML representation for engineering safety applications and a principled evaluation methodology.

The remainder of this paper is organized as follows. Section 2 provides background on functional modeling, DML, and the role of LLMs in fault diagnosis. Section 3 presents the research overview. Section 4 describes the proposed framework, including DML model construction, evaluation and interaction. Section 5 presents the case study. Section 6 reports results including evaluation and sample diagnostic insights. Section 7 discusses the findings, limitations, and future work. Section 8 concludes the paper.

# 2. Background

## 2.1. Functional Modeling and Dynamic Master Logic Model

Traditional diagnostic methods, such as ETA [1], [2], FTA [3], [4], and rule-based expert systems [16], [17], rely on modeling discrete events and predefined failure sequences. While effective for known fault-event combinations, these event-driven approaches are often incomplete for diagnostic purposes, as unanticipated sequences cannot be diagnosed, and complexity makes failure tracing impractical. In contrast, functional models represent a paradigm shift in reliability analysis, moving away from isolated event-based methods toward comprehensive, function-based evaluations. This shift is critical as system knowledge is inherently functional. Engineering systems are traditionally documented and conceptualized through their operational objectives, functional architectures, operating and mechanistic processes, rather than their associated failure modes and events. Event-based methods such as FTA and ETA represent a downstream abstraction of this functional knowledge and are often incomplete as a basis for causal reasoning. Functional modeling systematically captures system goals, functions, and their hierarchical dependencies rather than enumerating events. By representing expected functional behavior and component contributions, functional models enable fault detection through deviations from intended operation (e.g., loss of flow or information) instead of specific failure events [18]. The functional modeling approach improves completeness by supporting early-stage analysis, accommodating multi-fault and cascading scenarios through the purpose (functions) of constitutive elements of the system and any functional dependencies. By focusing on functional relationships rather than predefined event sequences, the approach can support multiple stages of the system life cycle, including design-phase FMEA, runtime diagnosis, and what-if analysis [19].

Building on these principles, DML provides a hierarchical framework for representing system objectives and goals, and the functions required to achieve them. Further, the model provides structural elements such as hardware operations, software executions, and human actions that realize those functions. The model is historically developed by applying the Goal Tree-Success Tree and Master Logic Diagram (GTST-MLD) construction rules [6]. In GTST-MLD, higher-level goals define why functions and structures are necessary, while lower-level functions describe how those goals and functions are achieved through interactions among hardware, software, and human elements. Logical relationships are represented explicitly through logic gate structures that encode functional dependencies, allowing structural composition and causal relationships to be modeled within a unified hierarchy [20]. The Dynamic Master Logic Diagram (DMLD) or simply DML extends the GTST-MLD foundation by incorporating time-dependent, uncertain, and evolving behaviors, supporting both static and dynamic reasoning for system diagnosis, failure analysis, and reliability assessment [7].

More recently, KG-DML has extended DML's capabilities through machine-readable representations that support computational reasoning and interaction while preserving the underlying DML structure. This progression from event-based FT/ET methods, through the functional hierarchy of GTST and GTST-MLD, to the dynamic reasoning of DML, and now to the machine-readable KG-DML, reflects a continuous effort to improve the completeness, scalability, and computational accessibility of functional system models.

Within the DML framework, two fundamental causal relationships can be inferred: the propagation of failures to their ultimate system-level consequences, and the identification of functional or structural dependencies necessary for successful subsystem operation. Figure 1 illustrates the conceptual DML structure, showing a hierarchical functional and structural hierarchy from system objectives to basic components. Causal ("Why-How") and compositional ("Part-of") relationships link high-level goals to low-level elements.

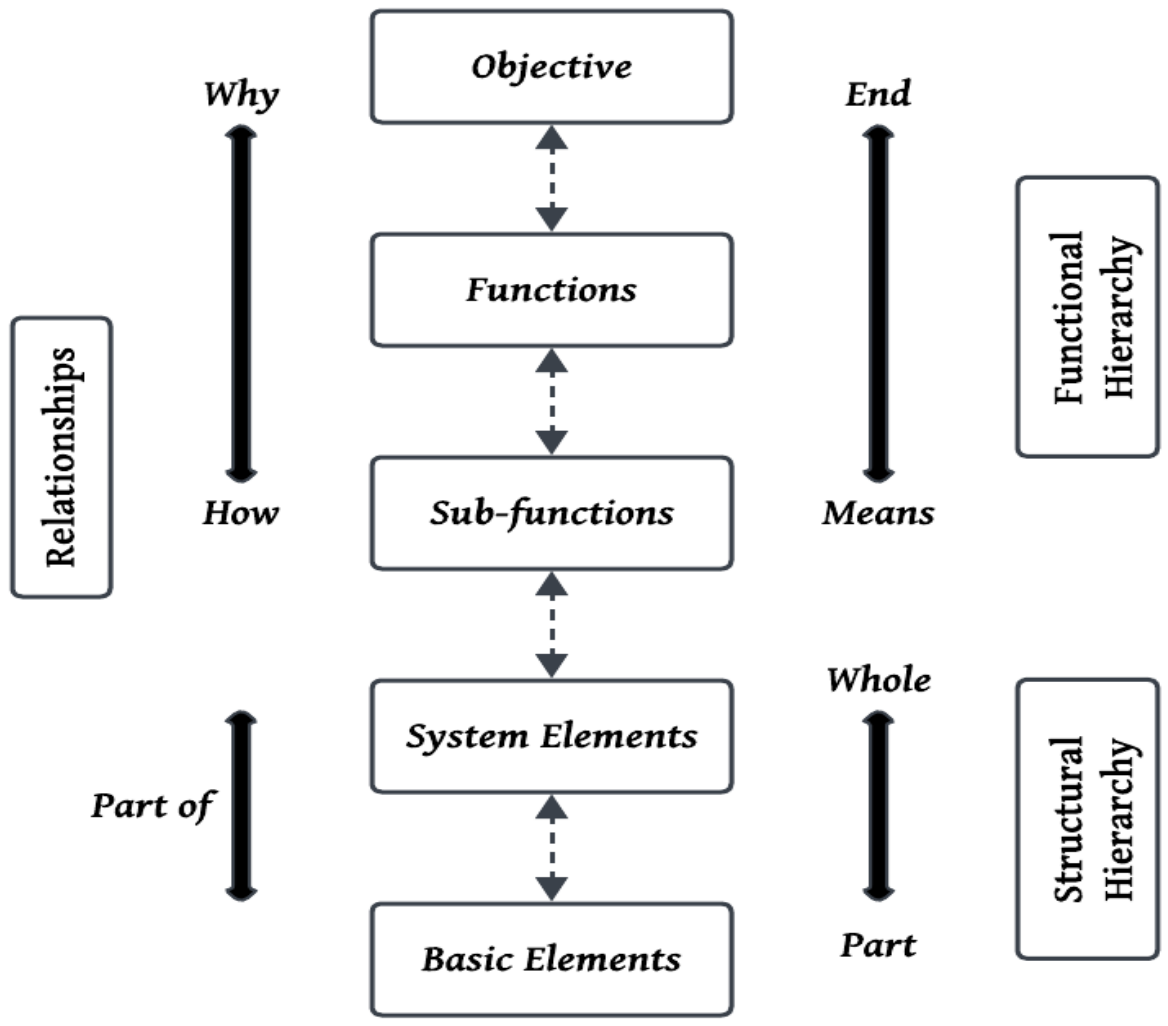

**Figure 1.** Conceptual DML framework [18].

The DML framework has been extensively applied across a broad range of domains. In nuclear and energy systems, these models have supported reliability and safety analysis, dynamic behavior modeling, and maintenance decision-making [21], [22], [23], [24]. Applications in aerospace and process supervision have focused on hazard identification, system interaction modeling, and real-time decision support during design and operation [25], [26]. In renewable energy and smart devices, the framework has enabled multi-state reliability assessment and fault tracing for systems such as wind turbines and intelligent sensors [27], [28], [29]. In software systems, DML has been applied to model safety-critical development processes, enabling requirement traceability and impact analysis across the lifecycle [30]. In business intelligence and expert systems, DML-based models have supported reasoning over tacit knowledge, uncertainty, and large-scale data streams for decision support and diagnostics [7], [31]. Related efforts have explored extensions of the DML framework for dynamic risk assessment by integrating the System-Theoretic Accident Model and Processes (STAMP) with modeling and simulation, enabling the analysis of complex interactions and time-dependent system behavior beyond traditional Probabilistic Risk Assessment (PRA) methods [32]. Throughout this paper, this family of models is collectively referred to as DML models.

While the hierarchical architecture illustrated in Figure 1 exhibits structural similarities to an ontology, the DML framework fundamentally diverges from conventional ontology-based knowledge representation. Traditional ontologies model static domain vocabulary, taxonomic classifications, and semantic entities. They are built on flexible relational graph structures in which concepts, properties, and semantic relationships may connect entities in any direction, with reasoning designed primarily for class-membership inference rather than functional system modeling. Conversely, the DML framework enforces a rigorous means-end hierarchy in which the decomposition flows strictly from the system-level objective down to subsystem and component functions and incorporates explicit logic that governs the operational conditions required to successfully realize the functions and attain the end objective. It is this combination of strict hierarchical decomposition and explicit success logic that makes DML suitable for intelligent reasoning about system behavior, functional dependencies, and failure propagation in ways that general-purpose ontology frameworks are not designed to support. In the present work, a KG is used as the underlying representation because its flexible graph structure can naturally accommodate the hierarchical decomposition, dependency relationships, and logical constructs required by the DML formalism. It is important to distinguish between the DML model and its KG representation: while DML defines the functional structure including objectives, functions, dependencies, and success logic, the KG serves only as the implementation and storage mechanism that encodes these constructs in machine-readable form, supporting storage, traversal, and reasoning without defining the functional structure itself.

## 2.2. LLMs Strengths and Limitations

Large language models have emerged as powerful tools for understanding and generating natural language across diverse application domains [33], [34]. Despite these capabilities, several limitations are relevant when considering their use in system modeling. Hallucination may occur when the available information is incomplete or when problem formulations are ambiguous, leading to the generation of outputs that are plausible but incorrect [35], [36]. To mitigate this issue, RAG methods incorporate external knowledge sources during inference, grounding model outputs in retrieved evidence and significantly reducing hallucination in knowledge-intensive tasks [37], [38]. In addition, performance can degrade for tasks that require consistent multi-step reasoning or explicit verification. This limitation has motivated the coupling of LLM-based agents with external tools, including prewritten code or functions, analytical solvers, databases, and rule-based systems, to provide computational grounding and validation in agent-based frameworks [39]. Another challenge is output variability across runs [40], particularly for complex tasks, which are commonly addressed through domain-specific fine-tuning [41] and structured prompt design to improve consistency and control [42]. This work does not aim to introduce new techniques to overcome these limitations, but rather it applies established mitigation strategies reported in the literature.

## 2.3. LLMs for Fault Diagnostics

Large language models have drawn increasing attention for fault diagnostics in industrial systems. While these studies demonstrate the growing application of LLMs in reliability and fault diagnosis, most focus on fault identification, classification, prognostics, or decision support using sensor, operational, and maintenance data. In contrast, the present work focuses on the automated construction of functional system models from technical documentation, a comparatively less explored application of LLMs within reliability engineering.

Fine-tuned LLMs have demonstrated strong diagnostic performance across complex systems and domain-specific applications [43], including near-perfect accuracy in HVAC diagnostics with robust generalization across unseen configurations [44]. Additional studies have extended LLM-based diagnosis to non-textual modalities, reformulating fault diagnosis using vibration signals [45], envelope spectrum images [46], and multi-task frameworks for joint fault diagnosis and remaining useful life prediction [47]. Hybrid architectures integrating real-time sensor streams with fine-tuned language models have further addressed scalability and deployment in industrial environments [48]. Agent-based frameworks have also been explored, enabling fault diagnosis directly from sensor data with human-readable explanations [49] and time-series anomaly detection [50].

To improve domain reasoning and interpretability, recent work has focused on integrating language models with structured knowledge representations. KG-RAG approaches have enhanced diagnostic reasoning and provided traceable reasoning paths in applications including CNC systems, railway bogie diagnostics, and high-voltage direct-current power systems [51], [52]. LLM-driven KG construction combined with retrieval-augmented reasoning has been demonstrated for cybersecurity KGs [53]. Related graph-based approaches have been applied to risk propagation analysis in railway systems [54] and to the generation of functional representations for battery energy storage risk assessment [55]. Beyond fault identification, RAG frameworks integrating fault knowledge bases and KGs have supported root cause analysis and maintenance recommendations in automotive domains [56], [57], while KG-integrated approaches have similarly been applied in software fault diagnosis and aviation assembly [58]. Systematic evaluation of LLM diagnostic outputs remains an open challenge, with recent work proposing evaluation methods using real-world fault logs to improve consistency and comparability [59].

Although these studies demonstrate the value of combining LLMs, KGs, and retrieval mechanisms, the resulting graph structures are generally used to organize domain knowledge, support information retrieval, or enhance diagnostic reasoning. To the best of our knowledge, limited attention has been given to the automated construction of executable functional models that preserve hierarchical functional dependencies and logical relationships required for diagnostic propagation. This distinction is particularly important for DML-based representations, where the model's structure directly determines the validity of diagnostic reasoning. Beyond supporting diagnostic reasoning, executable functional models preserve engineering knowledge in a structured representation that facilitates safety assessment, maintenance planning, and engineering decision-making. By enabling their scalable construction from engineering documentation, the proposed framework supports broader application of functional modeling to safety and resilience analyses in complex systems.

## 3. The Approach Overview

As discussed in Section 2, DML and KG-DML deliver functional representations that facilitate diagnostic reasoning via hierarchical dependency architectures. While our prior work [14] demonstrated the feasibility of KG-based DML representation, it did not address scalable construction from large, document-intensive system descriptions or systematic evaluation of structural completeness. The present work addresses these limitations through a retrieval-augmented construction framework that retrieves relevant portions of the system documentation at each stage of the DML hierarchy.

Since the constructed KG-DML serves as the foundation for diagnostic reasoning and system-level analysis, its structural correctness and completeness directly affect the reliability of downstream interaction. Accordingly, this study not only proposes a scalable construction method but also introduces an evaluation framework to assess model quality. Figure 2 presents a high-level view of the framework, including the model construction, evaluation, and interaction phases. During interaction, operational data (e.g., component states) may be incorporated as node attributes to support diagnostic reasoning. The resulting LLM-informed, KG-based approach supports the construction and interaction of DML models for diagnostic reasoning, reliability assessment, and decision-making in an engineering system. The proposed framework is designed to support the following key capabilities:

(a) Extracting functional elements and their dependencies from system documentation using retrieval-augmented LLMs, allowing functional decomposition from unstructured text.

(b) Layer-wise model construction for a complex system using targeted retrieval, allowing relevant portions of large-scale documentation to be selectively incorporated at each stage of the DML hierarchy.

(c) Systematic evaluation of the constructed KG-DML through layer-level precision and recall metrics combined with an integrity measure reflecting structural completeness and consistency.

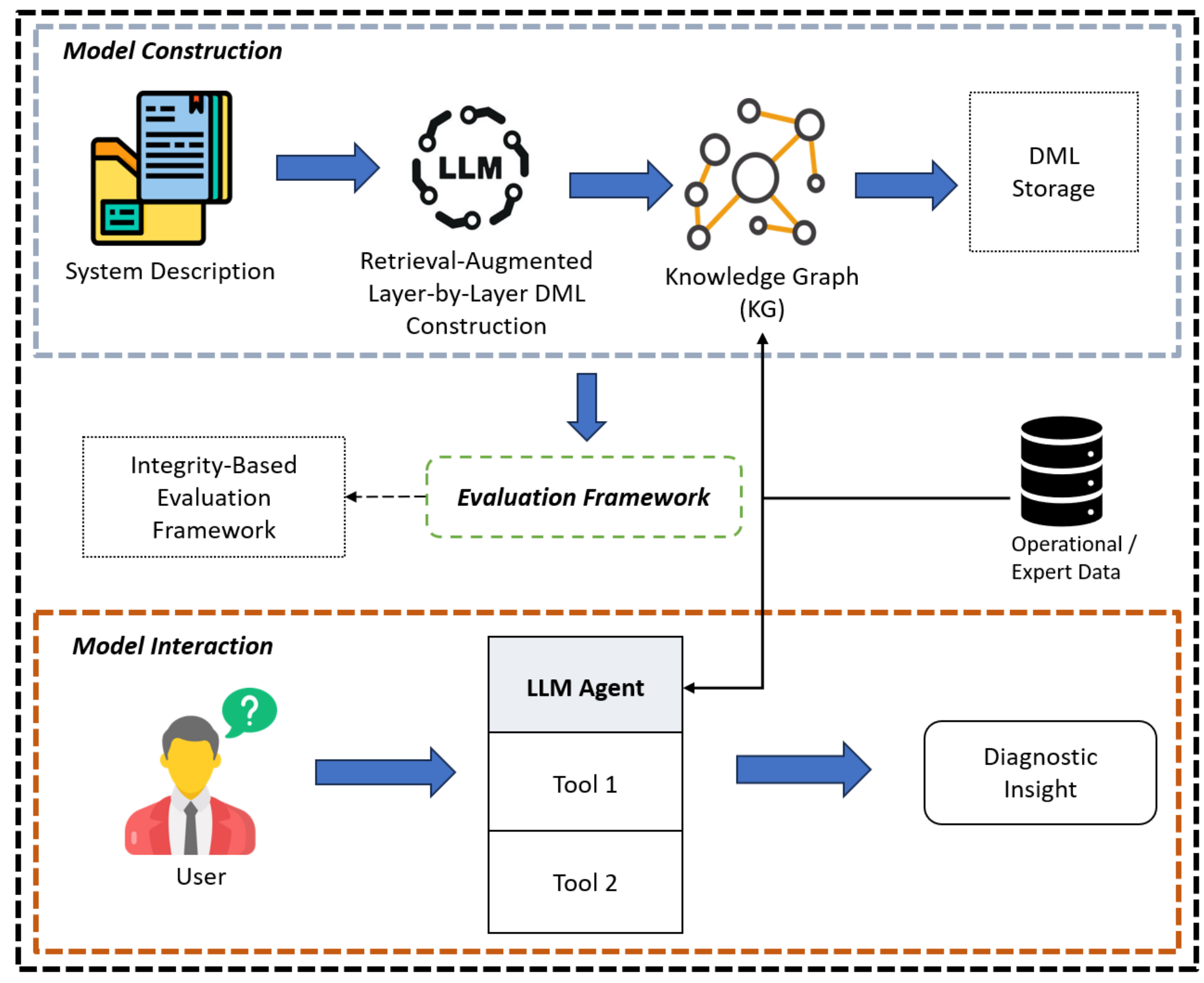

**Figure 2.** High-level overview of research. Adapted from [60].

A key concern in applying LLMs to KG construction is reliability, particularly when models are tasked with extracting arbitrary entities (e.g., persons, organizations, or locations) and relationships (e.g., employment, spatial, or part-whole relations, such as "works for," "located in," or "part of") from large, noisy, heterogeneous documents [61], [62], [63]. The framework presented here addresses this challenge by constraining the extraction process through a set of design choices that reduces ambiguity and limits the degrees of freedom available to the model. Rather than operating in an open-ended setting, the LLM is guided by a predefined schema, a structured template derived from the DML formalism, specifying node types, relationships, and hierarchical structure. In addition, the extraction task is restricted to the bounded scope of the system being modeled. Additionally, the LLM generates candidate elements and relationships, which are then verified and stored in a Java Script Object Notation (JSON) format. A separate module transforms this data into graph database queries and builds a connected graph that can be traversed. Together, these design choices shift the role of the LLM from unconstrained generation to structured interpretation within a controlled KG construction framework. The detailed implementation of these mechanisms is presented in Section 4.

## 4. Proposed Framework

This section presents the overall structure of the proposed framework. The primary contributions lie in the automated construction of the DML model from system documentation and in the development of an evaluation framework to assess the quality and integrity of the resulting KG-DML. In addition, the interaction module extends our prior work [14] by incorporating support for nested logical structures within the constructed model.

### 4.1. Model Construction

The process begins with preprocessing the system description to prepare the text for targeted retrieval during model construction. As shown in Figure 3, the documentation is first standardized to ensure consistent terminology and component identifiers. The text is then divided into overlapping segments and embedded in a vector store to support similarity-based retrieval of relevant passages. Model construction proceeds sequentially along the DML hierarchy, beginning with system goals and progressing through functions, subfunctions, components, and success conditions, where each success condition represents a specific performance indicator that defines whether a component fulfills its intended function.

At each layer, a retrieval query is constructed using the semantic definition of the current DML layer together with the parent elements identified in the previous stage. The most relevant portions of the system documentation are retrieved and supplied to the LLM, which generates candidate elements and their logical relationships in a structured format. The experiments were conducted using *GPT-4o* (OpenAI API), with the temperature parameter set to 0 to enforce deterministic model outputs. As validated elements are accepted, they are incorporated into a JSON representation that stores nodes and relationships in a machine-readable format. This master JSON representation evolves incrementally and serves as the record of all verified elements and relationships. For each subsequent layer, the parent nodes stored in the JSON are extracted and incorporated into the next retrieval query, thereby constraining evidence selection to documentation relevant to the already-constructed portion of the hierarchy.

After completion of all layers, the JSON representation is converted into Cypher, the declarative query language used by Neo4j [64] graph database used in this work, to generate the corresponding graph database statements. These statements are executed within the Neo4j environment to create the KG-DML representation. The resulting graph encodes the hierarchical decomposition and logical dependencies identified during construction and serves as the foundation for subsequent diagnostic reasoning.

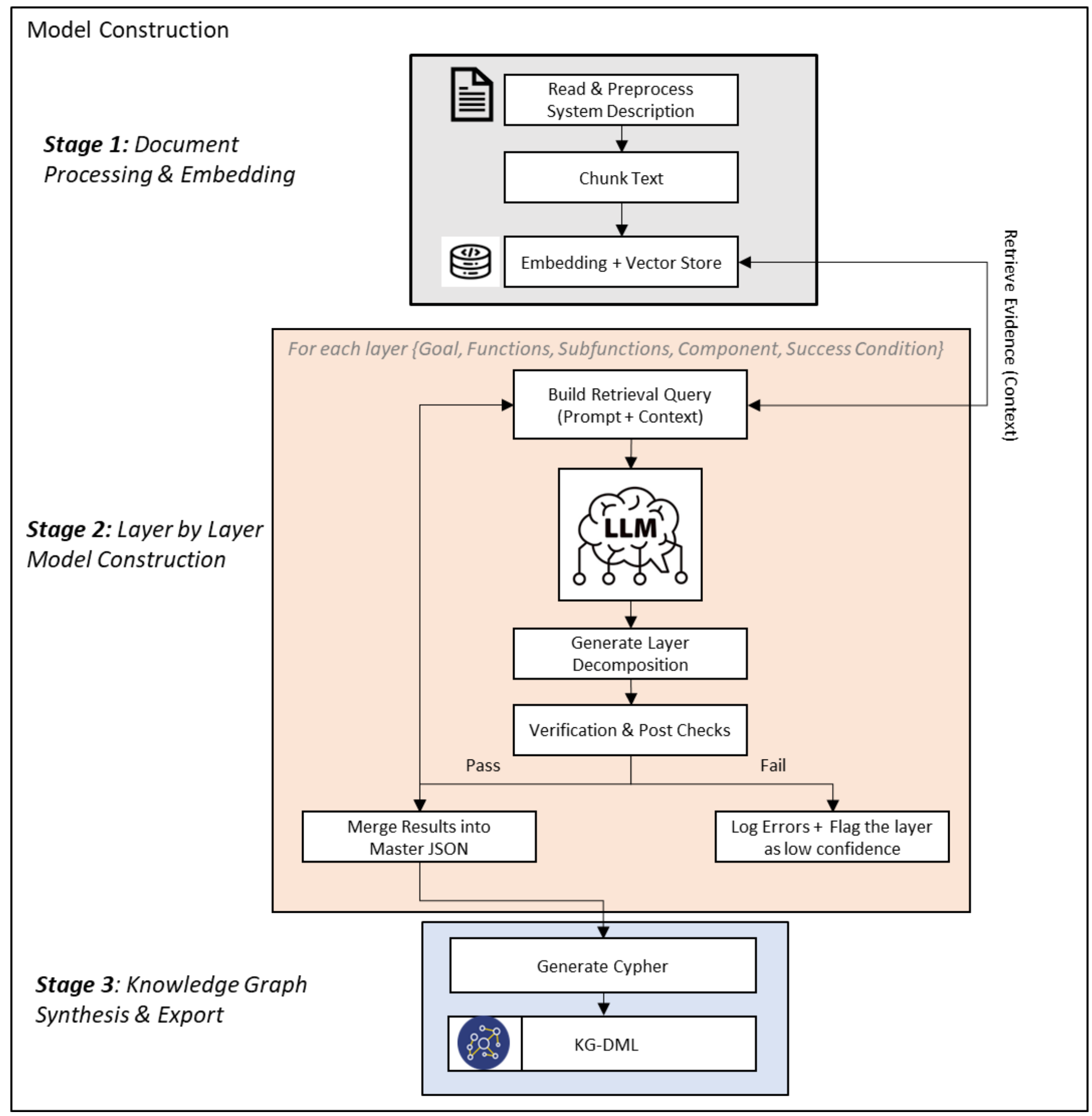

**Figure 3**. Overview of the workflow for constructing a DML model from system documentation.

### 4.1.1. Stage 1: Document Processing and Embedding

Prior to embedding and retrieval, the system documentation is preprocessed. This reduces ambiguity and supports reliable extraction. Component identifiers and tags are normalized to a consistent format to prevent inconsistent references to the same entity. Abbreviations are expanded and defined at first occurrence to maintain clarity throughout the document. When tagged components are first introduced, identifiers are supplemented with brief descriptive context, such as functional role or location, to associate each identifier with its meaning. This context is retained during embedding so that components with similar names, but different roles are less likely to be conflated during retrieval. Where possible, pronouns and shorthand references are resolved to explicit identifiers to maintain clarity across text segments. Together, these steps make retrieval more reliable and reduce the risk of confusing similarly named components by grounding each element in explicit identifiers instead of relying only on semantic similarity.

Following preprocessing, the documentation is segmented into text chunks using a paragraph-based strategy. Chunks are limited to 1,500 characters, with an overlap of 150 characters between adjacent chunks to preserve contextual continuity when information spans multiple paragraphs. Each chunk is embedded using the *text-embedding-3-small* model and stored in a vector database to support similarity-based retrieval during model construction.

### 4.1.2. Stage 2: Layer by Layer Model Construction

Stage 2 performs the structured construction of the DML model by progressing sequentially through the DML hierarchy. Rather than attempting to infer the entire model in a single step, each layer is constructed independently while being conditioned on the results of previous layers. Figure 4 illustrates this process for a representative layer. At the start of each layer, the current state of the model is stored in a master JSON representation. This master JSON serves as the intermediate model and contains all nodes and relationships that have been verified and accepted in earlier stages. Parent elements relevant to the current layer are read directly from this master representation and used to constrain the scope of the extraction task. To manage prompt length and control the scope of generation, parent elements at each layer are processed either individually or in small batches. A batch corresponds to a fixed number of parent elements that are processed together in a single model invocation. For example, a batch size of one process each parent element independently, while larger batch sizes allow multiple parent elements to be processed jointly. At a given layer, each parent element is expanded by generating its corresponding child nodes and the logical relationships that connect them within the DML hierarchy. The impact of batch size on model quality is examined in the Evaluation Section.

For each batch, a retrieval query is then constructed using two sources of information: the semantic definition of the DML layer being processed and the parent elements extracted from the master JSON. This query is embedded and compared against the embeddings of the preprocessed system documentation stored in the vector database. Similarity is measured using cosine similarity between the query embedding $q$ and each document chunk embedding $d_i$, defined as Eq. 1.

$$cos\,(q, d_i) = \frac{q\ \cdot d_i}{||q||\ ||d_i||} \qquad \text{Eq. 1}$$

In this expression, $q$ represents the embedding of the retrieval query constructed for the current DML layer, $d_i$ represents the embedding of the $i$-th text chunk from the system documentation, and $||\cdot||$ denotes the Euclidean norm used to normalize vector magnitude. The cosine similarity between the query embedding and each document chunk embedding is computed, and the top $K$ chunks with the highest similarity scores are selected as supporting evidence, where $K = 10$ in this work. The retrieved evidence, together with the parent elements and layer-specific instructions, is provided to the language model as a prompt. The model is tasked with identifying candidate elements at the current layer and defining their logical relationships to the parent elements, following the DML structure. Output is restricted to a predefined JSON schema to maintain consistency and enable automated verification. Once generated, the output is checked for structural correctness, identifier validity, and adherence to layer-specific constraints. Valid elements and links are appended to the master JSON. Outputs that fail these checks are logged, and the corresponding layer is flagged as low confidence for evaluation purposes. This procedure is repeated for each layer of the DML hierarchy until all have been processed.

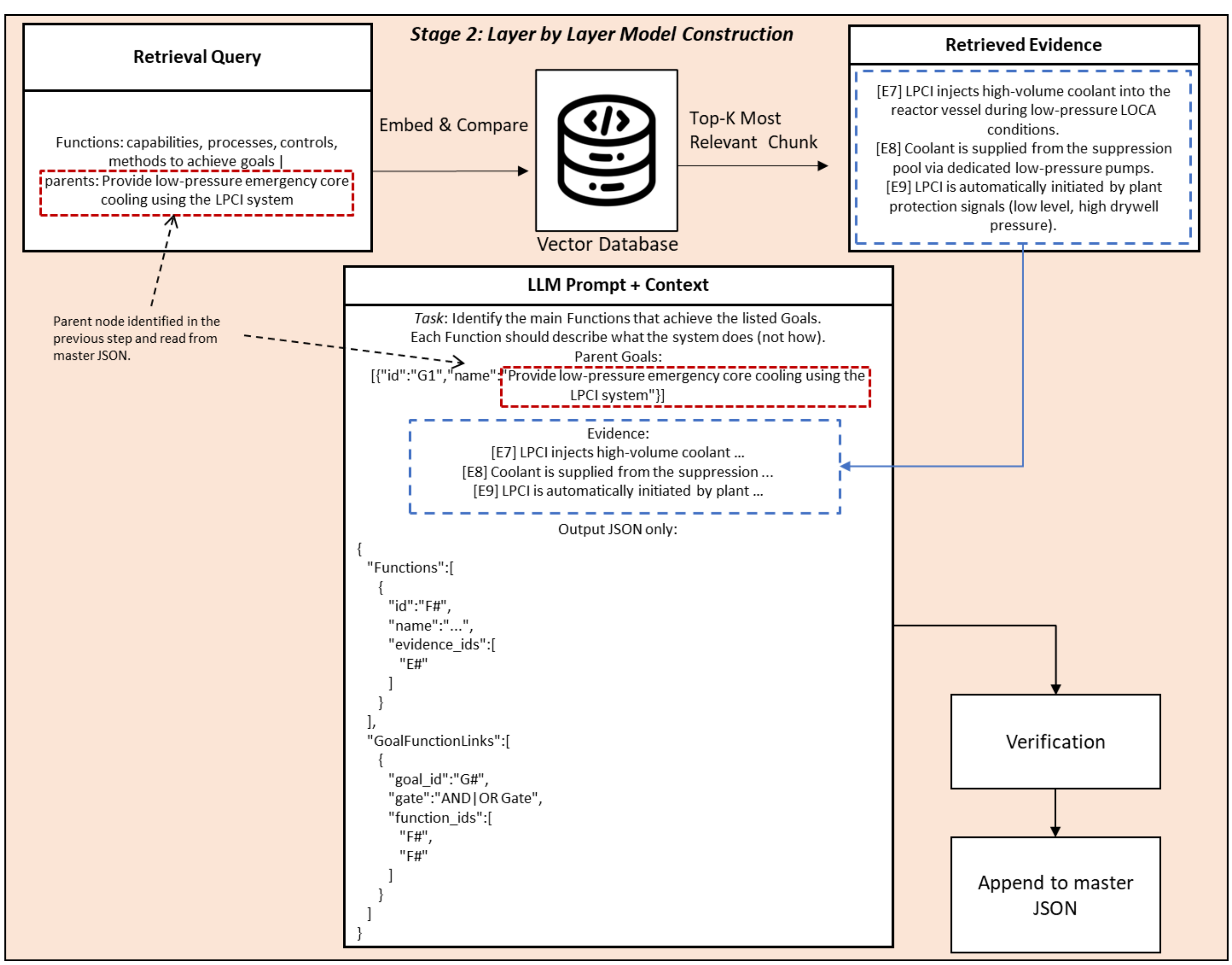

**Figure 4.** Layer by layer DML model construction process.

#### 4.1.3. Stage 3: Knowledge Graph Synthesis and Export

Stage 3 converts the validated master JSON produced in Stage 2 and exports it into a graph database. The master JSON contains the final set of accepted elements along with their parent-child relationships and any Boolean logic used to connect them.

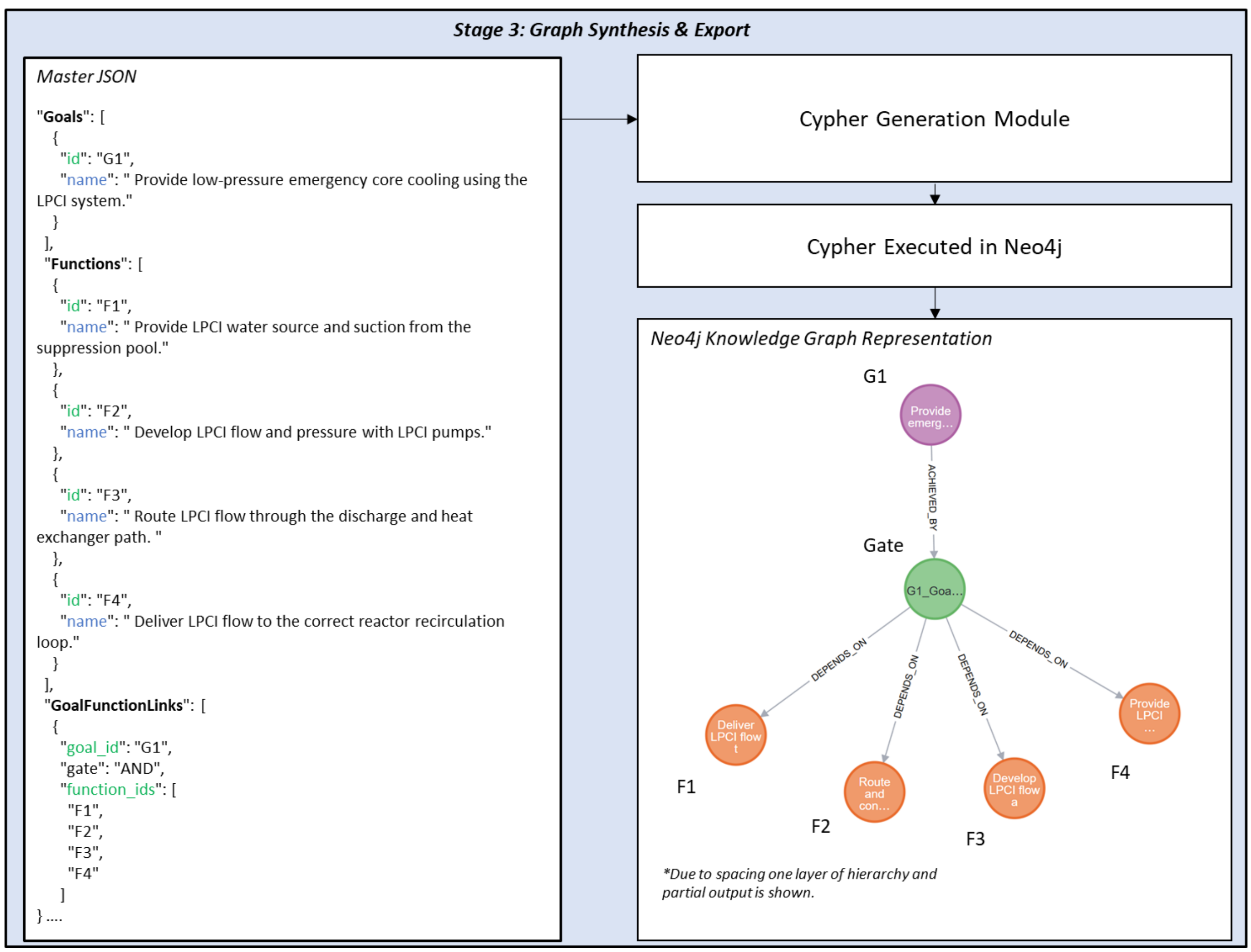

**Figure 5**. Graph Synthesis and Export Stage.

As shown in Figure 5, the master JSON is passed to a query generation module that translates the structured representation into executable graph database statements. The generation process follows Algorithm 1. Each node list in the master JSON is read to create the corresponding graph nodes with their identifiers and attributes, after which the relationship lists are processed to establish the parent-child links that define the DML hierarchy.

For relationships that include Boolean structure (AND/OR), the logical semantics are represented explicitly in the graph. When a dependency is governed by a gate, an intermediate gate node is created, connecting the parent element to its child nodes. If nested gates are present, the same procedure is applied recursively to preserve the multi-level logical structure without flattening the hierarchy.

**Algorithm 1.** Cypher Generation with Nested Boolean Gates

**Input:** JSON file *Master JSON file*
**Output:** Cypher script $kg.cypher$
1 Load JSON file into structure $D$
2 Initialize empty list $CypherLines$
3 **for all** $E \in \{Goal, Function, Subfunction, Component, SuccessCondition\}$ **do**
4 Add uniqueness constraint on $E.id$
5 **end for**
6 **for all** $G \in \{Goals, Functions, Subfunctions, Components, SuccessConditions\}$ **do**
7 **for all** $n \in D[G]$ **do**

8 Merge node with *id* = *n.id* and set *name*
9 **end for**
10 **end for**
11 **Procedure** $BUILDGATE(prefix, gate, parent)$
12 Create Boolean gate of type $gate.gate$
13 **if** $parent$ exists **then**
14 Create REQUIRES edge from $parent$
15 **end if**
16 **if** *gate* has $component_ids$ **then**
17 **for all** *c* **do**
18 Merge $Component\ c$ and connect via REQUIRES
19 **end for**
20 **end if**
21 **if** $gate$ has $subgates$ **then**
22 **for all** *sg* **do**
23 $BUILDGATE(prefix, sg, current)$
24 **end for**
25 **end if**
26 **End Procedure**
27 **for all** $l \in D[GoalFunctionLinks]$ **do**
28 Create gate and connect Goal via *ACHIEVED_BY*
29 **for all** $f \in l.function_ids$ **do**
30 Connect gate to Function via *DEPENDS_ON*
31 **end for**
32 **end for**
33 **for all** $l \in D[FunctionSubfunctionLinks]$ **do**
34 Create gate and connect Function via *DEPENDS_ON*
35 **for all** $sg \in l.subfunction_ids$ **do**
36 Connect gate to Subfunction via *DEPENDS_ON*
37 **end for**
38 **end for**
39 **for all** $l \in D[SubfunctionComponentLinks]$ **do**
40 $BUILDGATE(prefix, l, null)$
41 Connect Subfunction to root gate via REQUIRES
42 **end for**
43 **for all** $l \in D[ComponentSuccessLinks]$ **do**
44 Create gate and connect Component via *SUCCESS_THROUGH*
45 **for all** $s \in l.success_condition_ids$ **do**
46 Connect gate to $SuccessCondition$ via *REQUIRES*
47 **end for**
48 **end for**

## 4.2. Evaluation Framework

The objective of the evaluation is to assess the quality of the constructed DML model represented as KG-DML. Since the model is intended to support diagnostic reasoning through traversal, its usefulness depends not only on the correctness of individual elements, but also on the integrity of the relationships across the DML hierarchy. The evaluation framework therefore assesses element-level accuracy and system-level structural consistency by introducing an aggregate integrity score.

### 4.2.1. Model Normalization and Element-Level Evaluation

Before comparison, both the constructed KG-DML and the reference model are converted into a normalized representation. This step reduces differences that do not affect model meaning, such as variations in naming or identifier assignment. Element names are standardized using string normalization and semantic matching so that conceptually equivalent nodes can be aligned even when their textual labels differ slightly. This semantic alignment establishes a consistent basis for identifying matches, omissions, and incorrect inclusions across all layers of the model. The reference KG-DML model is visualized in Figure 6, where the total number of nodes at each hierarchical layer are indicated.

Following normalization, node-level evaluation is conducted independently for each layer of the DML hierarchy. For each layer, nodes in the constructed model are compared with the corresponding nodes in the reference model (referred to as the gold model). Each predicted node is classified into one of three categories: correctly identified, missing, or hallucinated. A node is considered correctly identified if a corresponding match is found in the gold model. Nodes that exist in the gold model but are absent from the constructed model are classified as missing. Conversely, nodes that appear in the constructed model but have no corresponding match in the gold model are classified as hallucinated.

These classifications are used to compute precision and recall at each layer. Precision reflects the proportion of identified nodes that are correct, while recall reflects the proportion of reference nodes that are successfully recovered. An $F_2$ score is also reported to place greater emphasis on recall, reflecting the fact that missing elements in a DML model are generally more detrimental than the inclusion of extra elements. Reporting these metrics separately for goals, functions, subfunctions, components, and success conditions provides insight into how extraction performance varies across different levels of abstraction.

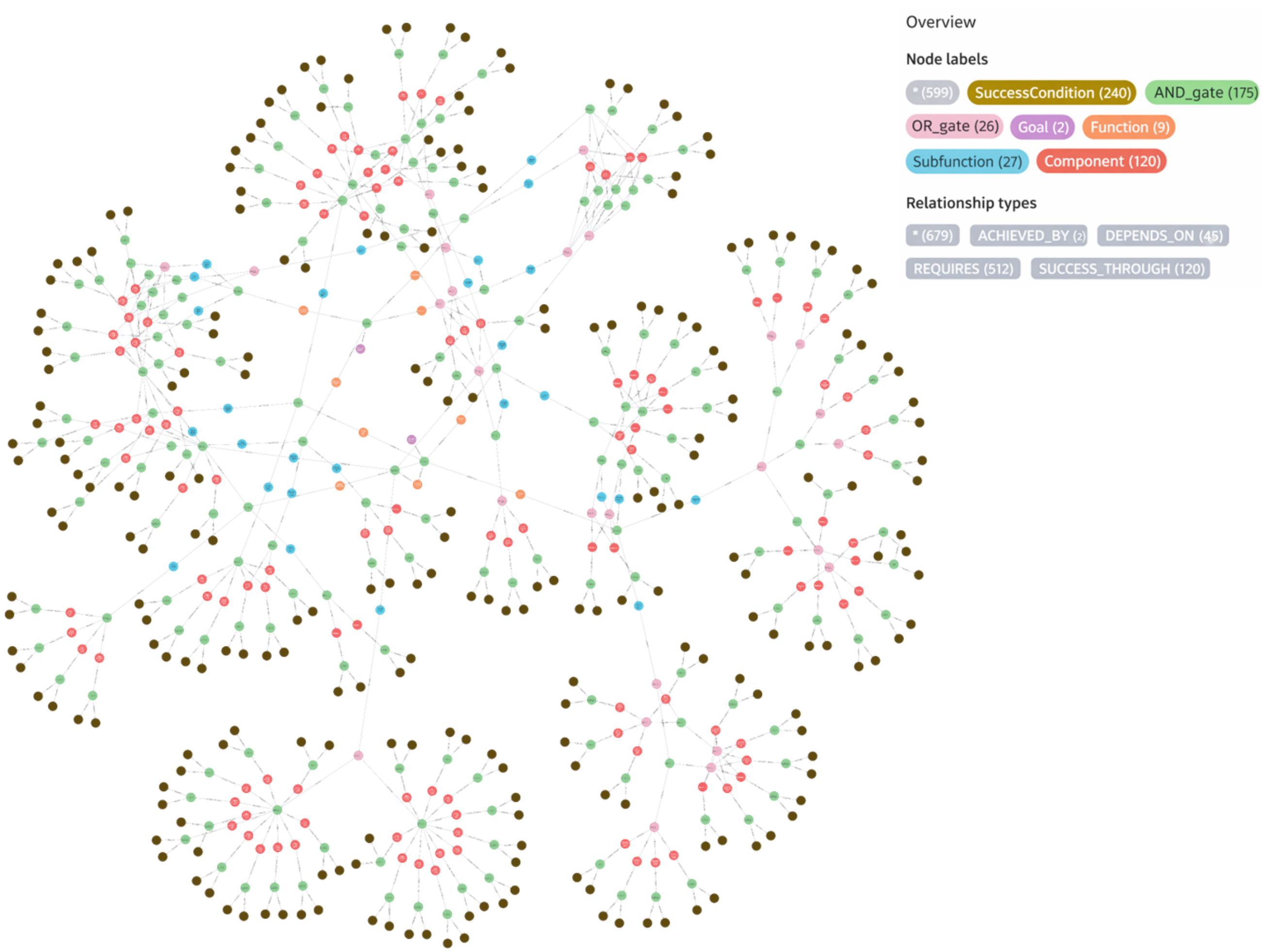

**Figure 6.** Reference KG-DML represented in Neo4j.

### 4.2.2. Relationship Structure and Logical Consistency

In addition to node identification, the evaluation examines whether relationships between nodes are correctly established. For each parent node in the reference model, the expected set of child nodes is compared with the set produced in the constructed KG-DML. Relationship errors are categorized as missing links, incorrect links, or extra links. Since logical relationships between parent and child nodes are represented through AND/OR gates at each hierarchical layer, gate accuracy is evaluated for all layers of the DML structure. Gate accuracy measures whether the logical operator connecting a parent node to its children matches the corresponding operator in the reference model. Gate correctness is evaluated separately from node and link identification, as an incorrect logical operator can alter how elements contribute to higher-level functions even when the correct nodes are present. Link evaluation considers only whether the correct parent-child associations are present, independent of logical structure. Different types of errors affect the usefulness of the KG-DML to different degrees. Errors occurring at higher levels of the hierarchy tend to propagate downstream and affect larger portions of the model. As a result, node omissions, relationship errors, and gate mismatches are weighted according to the DML layer at which they occur. Relationship and gate errors receive higher penalties when they disrupt complete functional paths from goals to success conditions.

### 4.2.3. Integrity Score Computation

Layer-specific precision, recall, and $F_2$ scores provide useful insight into the accuracy of individual elements within the DML hierarchy. However, these metrics do not capture the cumulative effect of errors across layers, nor do they reflect whether the constructed model preserves coherent functional paths from system goals to success conditions. To address this limitation, an overall integrity score is defined to evaluate the quality of the constructed KG-DML as a complete functional model, as shown in Eq. 2.

$$IntegrityScore = 100 * \left(\frac{\sum_l w_l * F_l}{\sum_l w_l}\right) * \exp\left(-\frac{E_{structural}}{(N_{nodes} + N_{links}) * S}\right) \qquad \text{Eq. 2}$$

where, $F_l$ denotes the $F_2$ score for layer $l$, $w_l$ is the importance weight assigned to that layer, and $E_{structural}$ represents the aggregate weighted structural penalty, computed from structural reconstruction errors including missing nodes, missing relationships, gate mismatches, and orphan nodes. The term $(N_{nodes} + N_{links})$ represents the size of the reference model, and $S$ is a penalty scaling factor that controls the sensitivity of the score to structural reconstruction errors. The exponential form of the structural penalty was selected to reflect the nonlinear impact of accumulated structural reconstruction errors on functional reasoning. An exponential decay function increases sensitivity to accumulated errors while preserving moderate penalties for isolated errors. This formulation allows the metric to distinguish between minor local discrepancies and structural reconstruction errors that affect end-to-end reasoning. Exponential formulations are commonly used to control how penalties accumulate in scoring functions, including exponentially weighted integrity or trust updates and exponential decay penalties in quality assessment metrics [65], [66]. The sensitivity of the integrity score to the penalty scale parameter $S$ is examined in Section 6 to assess the robustness of the metric with respect to this parameter. The first term of the equation computes a weighted average of the $F_2$ scores across all layers. This term reflects how accurately the framework identifies individual DML elements, accounting for both missing and extraneous elements, with an emphasis on recall. Higher-level layers are assigned larger weights to reflect their greater influence on downstream structure and reasoning. Layer weights were selected to reflect the relative impact of each DML layer on downstream reasoning and were held constant across all experiments.

The second term applies an exponential penalty based on the aggregate weighted structural reconstruction penalty. This penalty captures structural reconstruction errors, including missing nodes, missing relationships, gate mismatches, and orphan nodes. By normalizing the penalty with respect to the size of the reference model, the resulting integrity score remains comparable across systems of different sizes and levels of complexity. Missing and extraneous elements both influence the integrity score through the $F_2$ term, while structural reconstruction errors incur an additional penalty to reflect their impact on structural completeness and end-to-end functional reasoning.

Together, these terms yield a single integrity score that reflects both element-level accuracy and overall structural completeness. A score of 100 corresponds to a constructed KG-DML that fully matches the reference model in terms of nodes, relationships, and logical structure. Lower scores indicate increasing levels of structural deviation, either due to missing elements, incorrect connections, or disrupted functional paths. The integrity score is intended to complement layer-level metrics by providing an interpretable measure of how suitable the constructed model is for downstream diagnostic interaction and system-level analysis. Algorithm 2 provides the full computational procedure for integrity score evaluation, including node alignment, link comparison, gate verification, and orphan detection.

**Algorithm 2**. KG-DML Evaluation.

**Input:** $PredJSON$, $GoldJSON$
**Output:** $IntegrityScore$ and evaluation metrics
1 Load $PredJSON$ into structure $P$
2 Load $GoldJSON$ into structure $G$
3 Normalize node names and structural fields in $P$ and $G$
4 Initialize empty mappings $\text{Align}_l$ for each layer $L$
5 Initialize counters for node metrics and link metrics
6 Initialize $Penalty \leftarrow 0$
7 **for all** L $\in \{Goals, Functions, Subfunctions, Components, SuccessConditions\}$ **do**
8 Compute embeddings $E_L^p \leftarrow \mathcal{E}\ (\{name(p) \mid p\ \in \text{P}_\text{L}\})$, $E_L^G \leftarrow \mathcal{E}\ (\{name(g) \mid p\ \in \text{G}_\text{L}\})$
9 Compute similarity matrix $Sim_L$ between $E_L^p$ and $E_L^G$
10 Perform one-to-one matching using threshold $\tau_L$ to produce $Align_L$
11 Compute node $TP_L$, $FP_L$, $FN_L$ from $Align_L$
12 Compute $Precision_L$ $Recall_L$ and $F_{\beta,L}$
13 $Penalty \leftarrow Penalty + w_L * FN_L$
14 **for all** L $\in \{Goals, Functions, Subfunctions, Components, SuccessConditions\}$ **do**
15 Remap predicted node IDs in $P$ into gold ID space using $Align_L$
16 **for all** T $\in \{GoalFunctionLinks, FunctionSubfunctionLinks, SubfunctionComponentLinks,$
17 $ComponentSuccessLinks\}$ do
18 Convert predicted links $P[T]$ into edge set $E_P^T = \{(parent, child)\}$ in gold ID space
19 Convert gold links $G[T]$ into edge set $E_G^T = \{(parent, child)\}$
20 Compute edge $TP_T$, $FP_T$, $FN_T$ from $E_P^T$ vs. $E_G^T$
21 Compute $Precision_T$ , $Recall_T$, and $F_{\beta,T}$
22 $Penalty \leftarrow Penalty + w_T\ \cdot\ FN_T$
23 Compute gate accuracy for common parents (AND / OR comparison)
24 $Penalty \leftarrow Penalty + \lambda_{gate} *\ GateMismatch_T$
25 Identify orphan predicted nodes $Orphans\ =\ \{p\ \in\ P:$
$p\ is\ aligned\ but\ appears\ in\ no\ predicted\ aligned\ edge\}$
26 **for all** $o\ \in\ Orphans$ **do**
27 $Penalty\ \leftarrow\ Penalty\ +\ w_{layer}(o)$
28 Compute weighted mean of all $F_\beta$ scores:
$\bar{F}\ = \frac{\sum_i w_i * F_{\beta,i}}{\sum_i w_i}$

29 Compute final integrity score:

$$IntegrityScore = 100 * \bar{F} * \exp\left(-\frac{Penalty}{(N_{nodes}+N_{links})\,S}\right)$$

30 **return** $IntegrityScore$ and all intermediate metrics

## 4.3. Model Interaction

As shown in Figure 2, interaction with the constructed KG-DML is mediated by an LLM-based agent equipped with a set of predefined tools. These tools implement graph traversal and probabilistic reasoning procedures and are invoked by the LLM agent based on the user's inferred intent. This interaction paradigm follows the general approach introduced in our previous work [14], in which LLMs serve as a coordination layer that interprets user queries and selects appropriate model-based operations, rather than functioning as autonomous reasoning engines.

In addition to tool-based reasoning, the interaction framework also supports general questions about the model structure. Such queries are handled using a graph-augmented retrieval mechanism (Graph-RAG), in which relevant subgraphs and graph segments are retrieved and summarized without invoking the tools.

### 4.3.1. Upward Propagation

Upward propagation is a tool that evaluates system success and identifies impacted upper nodes by propagating success probability from the component level to higher-level functional and goal nodes. The probabilistic semantics and aggregation rules follow the formulation established in the prior framework. The primary difference introduced in the present work is the ability to evaluate nested Boolean gate structures, rather than restricting each dependency to a single flat gate. When a user poses a question such as "What is the likelihood that the system can achieve its objective under current conditions?" or "How does a specific failed component affect the overall goal?", the LLM agent invokes the upward propagation tool. The agent does not compute probabilities directly; instead, it selects and executes the tool which operates on the KG-DML.

The computation begins at the component and success condition level, where each component is associated with an operational state (e.g., working, degraded, failed) and a set of success conditions. The probabilities associated with component operational states and success conditions may be derived from operational data sources, including sensor measurements, system logs, inspections, or other observational evidence. These values are encoded into the KG-DML as node attributes and represent the current or assumed state of the system at the time of interaction. Each success condition defines conditional probabilities of satisfaction given the component's operational state. Component-level success probabilities are obtained by marginalizing over operational states and combining success conditions according to the Boolean logic defined for the component. These component success probabilities are then propagated upward through the hierarchy.

Since upward propagation operates directly on component state assignments, any dependencies among components are inherently reflected in these inputs. If multiple components are simultaneously degraded due to a shared cause, this condition is directly represented in their assigned states and propagated through the model. As a result, no additional qualitative augmentation is required in this mode, since dependency effects such as Common Cause Failures (CCFs) are already captured in the input conditions.

The upward propagation tool evaluates nested AND/OR gate trees recursively using standard Boolean logic, extending our earlier work to support sub-gates at any node in the hierarchy. Since the tool accepts simultaneous component state assignments, multi-fault scenarios are handled through the same propagation mechanism as single-fault cases, with the combined effect of multiple failed or degraded components evaluated directly through the Boolean gate structure.

### 4.3.2. Downward Propagation

In addition to diagnostic explanations, downward interaction also supports queries that ask how a specific objective, function, or subfunction can be achieved. The downward traversal logic follows the same structural principles as the previously established framework, with the extension in this work to accommodate nested Boolean sub-gates. When a user poses questions such as “How can this goal be achieved?” or “Generate the success paths for this function,” the LLM agent interprets the intent as a request for success path enumeration and invokes the corresponding downward traversal tool. In this mode, the tool generates minimal success path-sets by traversing the KG-DML from the queried node downward through its dependency structure. The traversal respects the Boolean logic encoded in the model, including nested (AND/OR) sub-gates. For conjunctive dependencies (AND), success paths are constructed by combining the required child elements, whereas for disjunctive dependencies, alternative paths are enumerated independently. At lower levels, paths are expressed in terms of components. The resulting success paths represent structurally valid combinations of elements that are sufficient to achieve the queried node.

In downward traversal, the generated success and failure paths are derived from the explicit Boolean logic structure of the KG-DML and therefore reflect structural dependencies among components. Common cause failure, however, may arise from both explicit structural dependencies and implicit coupling mechanisms, such as shared design, environment, maintenance practices, or underlying common root causes that are not explicitly represented in the model structure. Accordingly, when multiple identical or redundant components are identified within alternative success paths, the interaction layer augments the explanation by highlighting the potential for a CCF that could simultaneously affect these components. In addition to the success paths, the explanation therefore includes a qualitative indication that a shared cause could invalidate multiple paths concurrently. This augmentation does not modify the underlying logical structure but provides a partial representation of dependency effects beyond explicitly modeled relationships.

The treatment of CCF is partial, as it does not explicitly model the causal mechanisms or dependency structure within the KG-DML. Instead, it serves to flag the presence of potential shared vulnerabilities among components without altering the Boolean logic or path generation process. These paths provide insight into system design intent, redundancy, and functional alternatives, and can be used to support planning, comparison of design options, or what-if analysis.

### 4.3.3. Explanatory Queries

In addition to upward and downward propagation, the interaction framework supports explanatory queries that focus on the structural properties of the constructed KG-DML. These queries concern how elements are related within the hierarchy or request summary information about the model structure. For example, a user may ask how a specific function is related to a given subfunction, or how many total functions or components are defined in the model.

Such questions are handled through a Graph-RAG method in which the LLM agent generates a Cypher query based on the predefined schema of the KG-DML. The agent is provided with the node labels, relationship types, and hierarchical constraints defined during model construction, allowing it to produce syntactically valid and consistent queries. The generated Cypher statement is executed directly in the Neo4j database, and the resulting subgraph or aggregated output is returned to the LLM for interpretation and summarization into a natural language response. In this mode, the LLM does not perform reasoning over the structure directly. Instead, it acts as a query generator and response formatter, while the graph database performs structural retrieval.

## 5. Case Study

The case study uses documentation of the Low-Pressure Coolant Injection (LPCI) safety system used to protect the now decommissioned Millstone Point Unit 1 Boiling Water Reactor (BWR) in cases of loss of normal cooling. The documents used are from the Interim Reliability Evaluation Program: Analysis of the Millstone Point Unit 1 Nuclear Power Plant, Volume I prepared for the U.S. Nuclear Regulatory Commission [67]. Specifically, the LPCI system is intended to provide coolant to the reactor vessel following a loss-of-coolant accident (LOCA), ensuring adequate core cooling and preventing fuel damage under low-pressure conditions. The system is arranged as two redundant subsystems, designated A and B, which are interconnected by a normally locked-open crosstie line. Each subsystem includes multiple pumps, associated heat exchangers, spray headers, motor-operated valves, and defined injection lines. During operation, coolant is drawn from the suppression chamber and directed through alternative cooling and injection paths depending on system configuration and plant conditions. In addition to primary injection routes, the system incorporates torus spray lines, containment spray lines, and test lines, allowing for multiple flow configurations and operational modes. A simplified piping and instrumentation diagram (P&ID) illustrating the principal flow paths and subsystem interconnections is shown Figure 7. The presence of nested logical groupings and alternative functional paths makes the LPCI system a suitable case study for evaluating automated DML construction, particularly with respect to hierarchical decomposition, structural consistency, and the preservation of Boolean logic relationships.

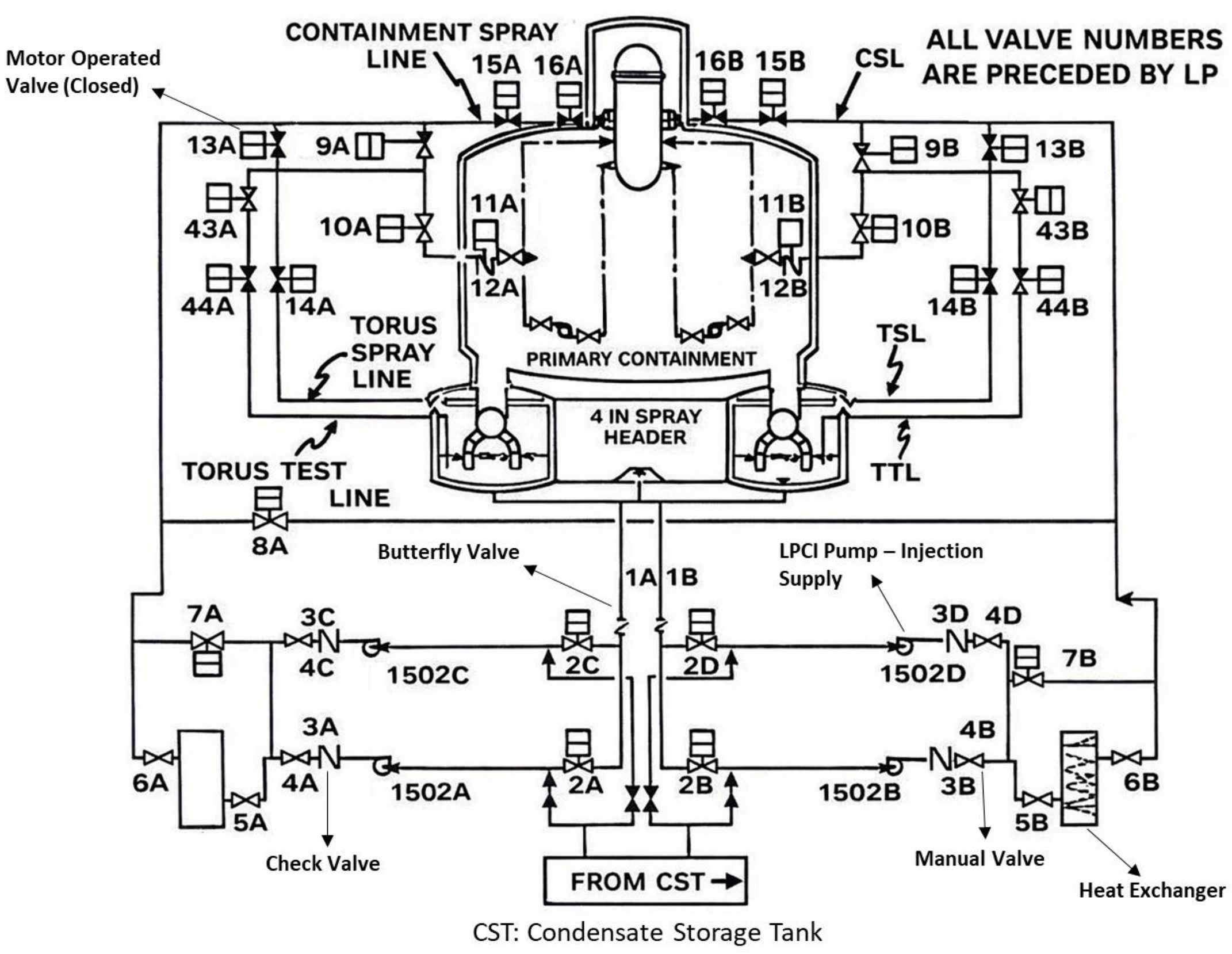

**Figure 7**. Simplified P&ID of the LPCI system [67].

## 6. Results

### 6.1. Evaluation Results

All quantitative results are reported across five independent executions of the model on the same system description and configuration. The metrics demonstrate consistent performance across runs, with low variability indicating stable behavior under the evaluated setting. Minor variations arise from residual nondeterminism in the pipeline, including differences in retrieval results and LLM inference, which may propagate through the hierarchical construction process. Full per-run values are provided in Appendix A to provide transparency and allow detailed inspection of where errors are concentrated, particularly at lower levels of the hierarchy.

At the node level, identification is perfect and consistent for the Goals, Functions, and Subfunctions layers, with precision, recall, and $F_2$ equal to 1.0 across all runs. Minor deviations from perfect score appear at the Component and Success Condition layers, where recall decreases slightly. For Components, $F_2$ ranges from 0.978 to 0.990, corresponding to approximately 1-3 missed nodes per run. For Success Conditions, $F_2$ ranges from 0.980 to 0.993, corresponding to approximately 2-6 missed nodes per run.

At the link level, higher-level relationships (Goal-Function and Function-Subfunction) are reconstructed perfectly across all runs. Variability is concentrated at the Subfunction-Component layer, where structural density is highest. At this level, recall ranges from 0.951 to 0.978, $F_2$ ranges from 0.961 to 0.982, and gate accuracy ranges from 0.889 to 0.963, corresponding to a small number of missing or misassigned links per run. These results reflect occasional mismatches in structural grouping and logical gate assignment. In contrast, the Component-Success Condition layer exhibits near-perfect performance, with $F_2$ ranging from 0.980 to 0.993, corresponding to approximately 1-2 missed links per run, and gate accuracy consistently equal to 1.0.

Table 1 reports the integrity score obtained under different batch size settings during layer-wise model construction. The highest score is achieved with a batch size of 1, with performance decreasing gradually as batch size increases. This trend indicates a trade-off between structural consistency and computational efficiency. Although individual runs show small fluctuations, the mean integrity score decreases consistently as the batch size increases (Table 1). The small variations across runs reflect the variability inherent in LLM-based generation and retrieval processes, where small differences in retrieved evidence or generated elements can propagate through the hierarchical construction pipeline. The largest batch size evaluated (15) represents a substantial prompt expansion in which multiple parent nodes are processed simultaneously. Within the examined range, the degradation in integrity score appears gradual rather than strongly nonlinear.

**Table 1.** Integrity score across different batch sizes.

| Run | Batch Size = 1 | Batch Size = 5 | Batch Size = 10 | Batch Size = 15 |
|---|---|---|---|---|
| 1 | 92.82 | 86.86 | 91.06 | 89.20 |
| 2 | 91.14 | 87.37 | 86.60 | 87.48 |
| 3 | 89.60 | 88.66 | 86.97 | 86.16 |
| 4 | 88.21 | 91.06 | 85.95 | 84.32 |
| 5 | 90.12 | 90.93 | 89.31 | 84.57 |
| **Mean** | 90.38 | 88.98 | 87.98 | 86.35 |
| **SD** | 1.73 | 1.96 | 2.14 | 2.04 |

Table 2 presents the sensitivity of the integrity score to the penalty scale parameter used in the exponential penalty term. Increasing the penalty scale leads to higher overall scores and reduced variability, while preserving the relative ranking across configurations. This behavior indicates that the integrity metric is sensitive to penalty scaling in magnitude but remains stable in relative comparison across configurations.

**Table 2.** Integrity score sensitivity to penalty scale parameter (Batch Size = 5).

| Run | Penalty Scale = 0.4 | Penalty Scale = 0.5 | Penalty Scale = 0.6 |
|---|---|---|---|
| 1 | 83.98 | 86.86 | 88.83 |
| 2 | 84.59 | 87.37 | 89.28 |
| 3 | 86.14 | 88.66 | 90.37 |
| 4 | 89.03 | 91.06 | 92.43 |
| 5 | 88.87 | 90.93 | 92.33 |
| **Mean** | 86.52 | 88.98 | 90.65 |
| **SD** | 2.35 | 1.96 | 1.68 |

## 6.2. Diagnostic Results

The diagnostic interface allows users to interact with the LPCI system model using natural language, supporting general and diagnostic queries. As shown in Figure 8, the interface supports downward propagation, upward propagation, and graph-based explanatory retrieval, depending on the nature of the user's request. Representative interaction examples are also illustrated in Figure 8.

For explanatory queries, such as how Subfunction SF3.2 (Provide proper flow through heat exchangers during injection mode) influences Function F3 (Route and condition LPCI flow through the discharge and heat exchanger path), the system retrieves and summarizes the relevant structural relationships directly from the graph. The results show that SF3.2 ensures that LPCI flow passes through at least one operational heat exchanger leg, which is required for F3 to maintain a properly conditioned discharge path. This response is grounded in the explicit parent-child dependencies encoded in the KG-DML.

When the user queries about the effect of the simultaneous failure of the heat exchanger inlet and outlet valves, the system identifies the resulting impact across the hierarchy. The disruption first affects Subfunction SF3.2, then propagates to Function F3, and ultimately impacts Goal G1 (Provide emergency low-pressure coolant injection to prevent core damage during a LOCA). The output traces the functional consequence from component-level failure to the system-level objective.

When queried about how Subfunction SF1.3, which provides suction flow path alignment, can be achieved, the system returns alternative success paths within the modeled hierarchy. The results identify valve configurations that establish a valid suction routing path, where each path corresponds to a combination of a manual suction shutoff valve and associated motor-operated suction valves required to complete the flow route, reflecting the Boolean dependency structure encoded in the KG-DML. For subfunctions supported by redundant identical components, the system returns alternative success paths, while the explanation layer highlights the potential for CCF due to shared design or operating conditions.

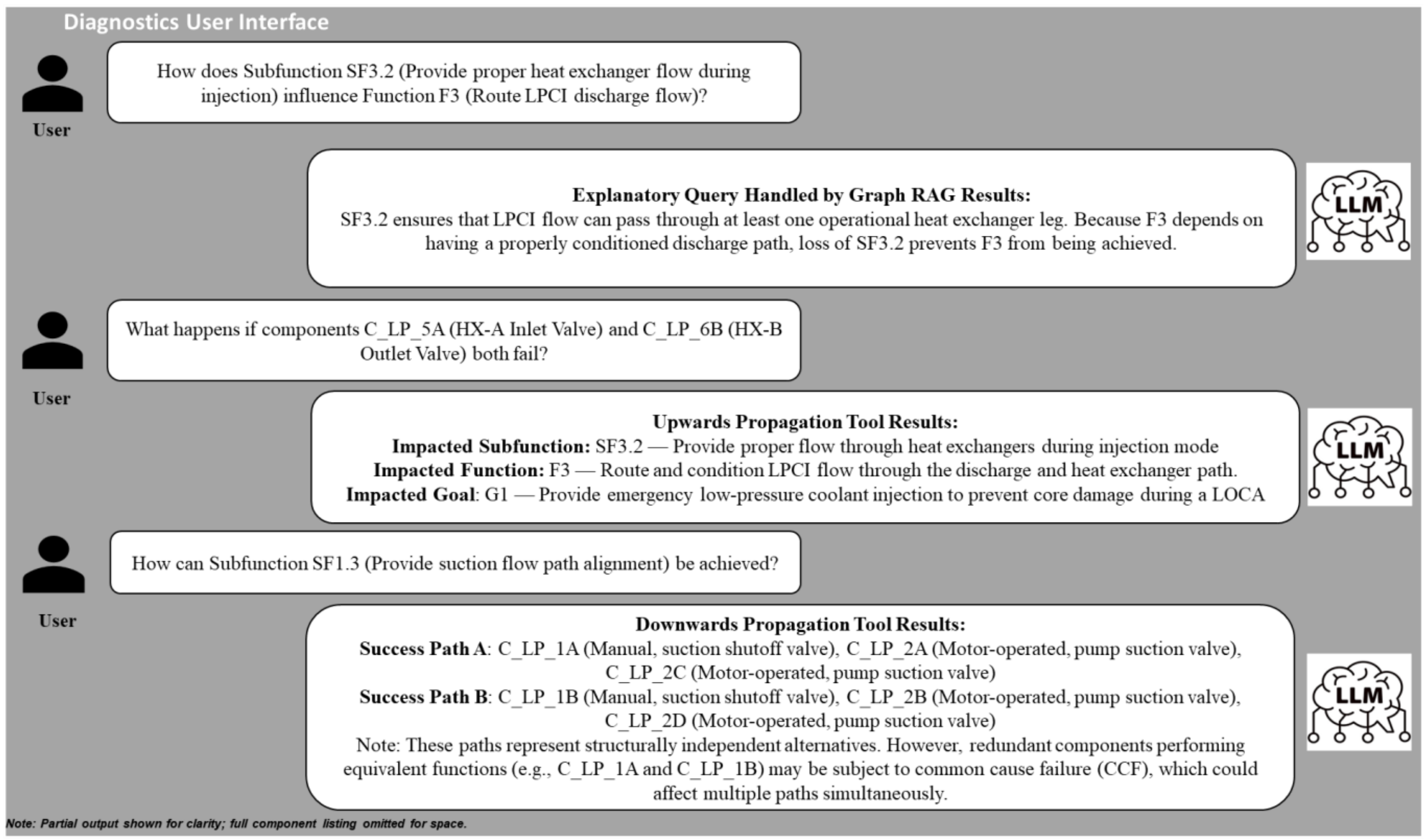

**Figure 8.** Interaction interface example containing user sample questions.

## 7. Discussion

The proposed LLM-KG-based diagnostic framework supports automated extraction and structuring of functional models from system documentation while maintaining high structural accuracy relative to the reference model. This approach has the potential to reduce the manual effort required for constructing DML models compared with traditional expert-driven development processes. In conventional practice, constructing a DML model for a complex engineered system often requires several months of manual effort. Domain experts must review extensive technical documentation, identify system goals and functions, extract component relationships, and assemble these elements into a coherent hierarchical structure. The framework presented in this study streamlines this process by leveraging LLM-assisted extraction and KG representation, reducing model construction time to a matter of days for complex systems while preserving logical consistency and diagnostic capability.

Small deviations from perfect performance are observed at the Component and Success Condition layers. Although precision remains very high for both layers, recall decreases slightly, leading to modest reductions in the $F_2$ score. Since $F_2$ places greater emphasis on recall, this pattern reflects occasional omissions at the more detailed levels of the hierarchy. Lower-level elements are more numerous and more closely tied to specific contextual descriptions, making them more sensitive to retrieval scope and batching decisions. Nevertheless, the magnitude of variation is limited, and overall performance at these layers remains high.

The link-level evaluation further confirms the structural fidelity of the constructed models. For the Goal-Function and Function-Subfunction relationships, both structural associations and logical operators were identified correctly across all runs. This suggests that once higher-level elements are extracted, their hierarchical dependencies are consistently reconstructed. Greater variability is observed at the Subfunction-Component layer. While structural precision remains high, recall is slightly reduced, and gate accuracy exhibits measurable variation. This layer represents the point in the hierarchy where structural density and nested logical relationships increase, which explains the observed sensitivity. In contrast, the Component-Success Condition relationships show near-perfect structural and logical consistency, indicating the stable recovery of these dependencies once components are identified.

The analysis of batch size highlights an important practical consideration. The highest mean integrity score was obtained with a batch size of 1, and performance decreases gradually as the batch size increases. The gradual decline in mean integrity score with increasing batch size suggests that expanding multiple parent nodes simultaneously increases prompt complexity and may introduce competition for context within the LLM input. This can lead to minor omissions or inconsistencies during hierarchical extraction. Larger batch sizes require the model to expand multiple parent elements simultaneously, increasing prompt complexity and the potential for minor inconsistencies. Smaller batch sizes isolate each expansion step, which appears to promote greater structural stability. Although a batch size of 1 produced the highest mean integrity score, the improvement relative to a batch size of 5 was limited. Reducing the batch size from 5 to 1 increased the number of LLM invocations approximately fivefold, from roughly 32 to 158 calls across all construction layers. However, since smaller batch sizes process fewer parent elements per call, each individual prompt is shorter and requires fewer output tokens, which partially offsets the increase in total execution time. Nevertheless, the cumulative overhead in API usage and execution time remains substantial relative to the marginal gain in structural accuracy. A batch size of 5 provided a balance between structural consistency and computational efficiency, maintaining high integrity scores while reducing both execution time and API call volume.

It should be noted that this analysis focuses solely on computational cost and model performance. The potential reduction in human modeling effort enabled by automated DML construction is not explicitly quantified here, as the evaluation does not incorporate human labor costs. In practical applications, such human cost considerations may further influence the choice of configuration.

The sensitivity analysis on the penalty scale parameter shows that the integrity score responds in a predictable manner to changes in the exponential penalty formulation. As the penalty scale increases, the overall score increases and variability decreases. This behavior is consistent with the intended role of the parameter, which modulates the impact of structural omissions. The relative ordering of results remains stable across parameter values, indicating that the metric behaves consistently within the examined range.

Overall, the results indicate that the framework produces a structurally coherent and reproducible DML representation. It should be noted that the Boolean structure of the KG-DML assumes independence among components when generating success paths. In practice, CCF may introduce dependencies among identical components, or collocated components; in this work, such effects are not structurally modeled but are instead communicated through qualitative annotations during interaction. Variability across repeated runs is limited, and deviations are concentrated in the most structurally dense portion of the hierarchy. The evaluation also demonstrates that the integrity score provides a stable aggregate measure of model quality while remaining sensitive to batching choices and penalty scaling decisions. Although the framework accelerates DML construction, it is intended to support rather than replace expert modeling practice. The final review and validation of the generated model remain the responsibility of domain experts, particularly for detailed lower-level elements.

The principal contribution of the proposed framework in this paper lies in reducing manual extraction and structuring effort while preserving expert oversight and control. The high structural accuracy and low variability observed across runs can be attributed in part to the constrained design of the extraction process. By limiting the degrees of freedom of the LLM through schema grounding, bounded scope, and structured generation, the framework reduces ambiguity and mitigates common failure modes associated with unconstrained KG construction.

Although the proposed framework is demonstrated on a single system, it can also be applied at the subsystem level. In practice, large systems are usually broken down into smaller parts, and the same approach could be used to construct KG-DML models for each of these subsystems separately. The contribution of this work is in providing a consistent way to build these models from documentation. These subsystem models could then be combined into a master model representing the full system, although additional work would be needed to handle dependencies across subsystems.

### 7.1. Limitations

Several limitations of the present study should be acknowledged. First, the evaluation is conducted on a single case study system. Although the selected system is structurally nontrivial and representative of safety-critical engineered systems, the generalizability of the framework to substantially larger or more complex systems has not yet been examined. In particular, model behavior under increased documentation volume distributed across multiple documents, higher component density, or more nested logical structures may introduce additional variability that is not fully reflected in the present results.

Second, the evaluation framework emphasizes structural consistency at the node, link, and logical gate levels. While this approach enables systematic assessment of hierarchical alignment, it does not directly evaluate functional equivalence. Structural similarity does not necessarily guarantee identical logical behavior during upward or downward propagation. A comparison based on minimal cut sets or success path sets would provide a stronger test of behavioral equivalence. However, even complete cut set and path set equivalence would not constitute full functional validation of a DML-based model, as these metrics operate at the structural component level and do not capture the hierarchical decomposition of objectives into functions and subfunctions or how failures propagate through that functional hierarchy. Expert validation of the functional layer would therefore remain a necessary complement to cut set and path set comparison. Cut set and path set comparison was not included in this study and remains an important direction for future work.

The present study evaluates the overall framework rather than the contribution of individual implementation components. A no-retrieval baseline was not evaluated, as full-context extraction is not feasible for the system documentation used in this study: its volume substantially exceeds practical context-window and generation limits, making retrieval a structural requirement of the framework rather than a configurable design choice to be benchmarked against an alternative. Future work should include systematic ablation studies and baseline comparisons examining the effects of retrieval augmentation, layer-wise construction, prompting strategies, model selection, and parameter settings. Such analyses would provide a more detailed understanding of the factors influencing KG-DML construction quality and framework performance.

It is also worth noting that this evaluation gap is not equivalent to the ablation practices common in the general LLM-based KG construction literature, where models extract arbitrary entities and relationships from unstructured text without a predefined structural target and can therefore be meaningfully compared against unconstrained or alternative extraction baselines. The present framework instead targets a specific, schema-constrained functional representation, in which the LLM operates within a predefined DML template rather than performing open-ended extraction. Because the schema, hierarchy, and logical structure are fixed rather than learned or selected among alternatives, standard ablation designs from the open-domain KG literature do not directly transfer to this setting, and a dedicated ablation methodology tailored to schema-constrained, hierarchical construction tasks remains an open direction for future work.

The framework also depends on retrieval configuration and model behavior for information selection and interpretation. Since model construction is conditioned on embedding-based similarity and top-k evidence retrieval, variations in chunk segmentation, retrieval ranking, or similarity thresholds may alter the contextual evidence supplied to the language model. In addition, the framework was evaluated using a single large language model configuration. Performance may vary across alternative model architectures or capability levels, and robustness across models was not systematically examined. Model outputs may also be sensitive to prompt formulation and instruction structure, which were held fixed in this study.

The integrity score introduces additional design choices that influence aggregate evaluation results. Although sensitivity to the penalty scale parameter was examined, the selection of layer weights and the structure of the penalty term reflect modeling decisions. Alternative weighting schemes or penalty formulations could produce different numerical scores even when underlying structural discrepancies are comparable. The current implementation also assumes a largely static representation of the system.

The proposed framework does not explicitly represent CCF mechanisms within the KG-DML structure. Dependencies arising from potential CCFs among identical or coupled components are instead addressed qualitatively during the explanation stage. Consequently, the current framework does not capture shared failure mechanisms or dependency strengths required for quantitative CCF analysis.

While the DML model supports dynamic reasoning once constructed, the extraction process is based on static documentation and therefore yields a snapshot of the system structure. Time-dependent or context-dependent behaviors are not explicitly modeled in the present framework, though such dynamics could be incorporated through extensions such as time-dependent delay gate formulations.

Finally, although the framework is evaluated in terms of the structural accuracy of the constructed KG-DML (e.g., node, link, and logical consistency metrics), reductions in human modeling effort are not formally quantified in this study. The proposed framework automates portions of the model construction process, but the extent to which this reduces expert labor has not been systematically measured. While the framework appears to reduce DML model construction time from several months of expert effort to a few days in the present case study, this estimate is based on limited practical experience rather than a controlled comparison. A systematic evaluation comparing manual and automated DML construction, including metrics for expert time and cost, would provide a more rigorous assessment of the framework's practical impact. Although preprocessing and normalization are used to mitigate ambiguity, the present study does not systematically evaluate performance under noisy, incomplete, or highly inconsistent documentation. Assessing robustness under such conditions remains an important direction for future work.

### 7.2. Future Work

In practice, system information is distributed across design specifications, operating procedures, maintenance manuals, and other technical documentation. Applying the framework to multi-source and multi-format documentation would allow a systematic assessment of robustness to inconsistencies, redundancy, terminology variation, and document fragmentation. In realistic engineering environments, relevant evidence is often distributed across independent documents rather than contained within a single source. Extending the framework to support RAG across distributed document sets would enable coordinated evidence selection from multiple sources during hierarchical model construction. Such extensions would require structured cross-document retrieval, version reconciliation, and mechanisms for resolving conflicting or overlapping descriptions. While the present case study demonstrates the approach on a substantially larger system than previously considered, additional validation across larger-scale engineered systems and distributed documentation environments is needed to more fully assess scalability and generalizability.

Another area for future work is modular construction across subsystems. Instead of building a single KG-DML representation at once, one could develop individual subsystem models and later integrate them into a comprehensive model of the entire system. This would require methods to effectively integrate subsystem models and manage dependencies across their interfaces.

Future work should investigate the explicit representation and quantification of CCF within the KG-DML framework. While the current approach can highlight potential shared vulnerabilities among redundant components, it does not model common cause dependencies within the underlying logic structure. Quantitative CCF representations could be incorporated using established approaches such as explicit beta-factor modelling for multiple CCF groups [68], as well as alpha-factor models [69]. Such extensions would enable explicit representation of shared-cause dependencies and support more rigorous reliability analysis and probabilistic reasoning within the KG-DML framework.

Beyond structural validation, future work should incorporate functional equivalence testing. While the present study evaluates node-, link-, and gate-level consistency, a stronger validation would compare minimal cut sets or success path sets between constructed and reference models. This would provide a direct assessment of whether the KG-DML reproduces equivalent logical behavior under upward and downward propagation, rather than relying solely on structural correspondence.

While the proposed KG-DML framework is not based on stochastic state-transition modeling, certain conceptual parallels may be drawn with Markov-style approaches, particularly in terms of directional dependencies and propagation behavior. Future work may explore whether insights from probabilistic transition models can inform extensions of the framework for dynamic or uncertainty-aware analysis. Future extensions may incorporate explicit modeling of CCF by identifying coupling mechanisms among components, such as shared design features, co-location, maintenance dependencies, or exposure to common environmental conditions. Retrieval-augmented LLM methods could be used to infer such relationships from system documentation by detecting components subject to the same root causes, enabling a more complete representation of dependency effects within the KG-DML.

Enhancements to the retrieval strategy, such as adaptive chunking, hybrid keyword and embedding retrieval, or domain-tuned embedding models, could reduce sensitivity in densely connected layers such as Subfunction-to-Component relationships. Systematic comparison of retrieval configurations would clarify robustness boundaries and support the development of configuration guidelines for larger or more complex systems. Further, alternative model architectures could reduce reliance on structured preprocessing. Multi-model pipelines, in which one model performs identifier extraction and normalization, and another model focuses on hierarchical decomposition and logical structuring, may allow more direct ingestion of heterogeneous documentation formats with less manual normalization. Comparative evaluation across different language model architectures would also clarify the framework's robustness to model selection. Similarly, ablation studies comparing hierarchical retrieval with and without parent-element conditioning would clarify the contribution of parent conditioning to retrieval quality and downstream construction accuracy.

The current implementation produces a largely static structural model derived from documentation. Future extensions could incorporate time-dependent or delay gate formulations to represent dynamic behaviors more explicitly. Such extensions would enable modeling of sequencing effects, temporal dependencies, and context-dependent system states within the DML structure. In addition to upward and downward propagation tools, the interaction framework could be extended to support more advanced reliability and importance analyses. For example, component importance measures, such as Birnbaum or criticality importance, could be incorporated to quantify the relative influence of individual components on system-level objectives [70]. Consistent with the functional formulation of DML, these measures would be evaluated in the success space rather than the failure space.

An additional practical consideration concerns model size and deployment constraints. The present implementation relies on a high-capability language model accessed through an external API to ensure stable and accurate hierarchical extraction. While this configuration performed well in our experiments, it may not always be the most cost-effective or practical option for large-scale or routine industrial use. Exploring hybrid configurations, in which lightweight models handle well-defined extraction steps and larger models are reserved for more complex reasoning tasks, could reduce computational cost while maintaining structural quality.

Finally, engineering system documentation is often proprietary, sensitive, or subject to regulatory restrictions. In such settings, transmitting system descriptions to external APIs may not be feasible. Future work should therefore investigate locally hosted language model deployments that allow confidential documentation to remain within secure organizational boundaries. Systematic evaluation of how model size, deployment architecture, and data governance constraints affect structural accuracy and integrity score performance would provide practical guidance for secure and cost-conscious adoption of the framework.

## 8. Conclusions

This study introduced a framework for constructing DML models directly from system documentation and representing them as a KG-DML structure to support diagnostic reasoning. By combining layer-by-layer retrieval with structured generation and graph synthesis, the approach converts unstructured technical text into a functional hierarchy that can be stored, queried, and analyzed computationally. The objective was not to replace expert modeling, but to reduce the manual effort required to extract and organize system structure from documentation.

Application of the proposed approach to the LPCI safety system of a boiling water reactor case study showed that the overall functional structure can be reconstructed consistently across repeated executions. Higher-level goals and functions were identified without variation, while minor differences appeared at the Component and Subfunction-Component layers, where structural density and logical combinations are greater. The evaluation framework, including layer-level metrics and an integrity score that accounts for structural omissions, provided a systematic way to assess model completeness and logical consistency. Sensitivity analysis highlighted the trade-off between batching configuration and structural stability.

The resulting KG-DML representation supports three types of interaction: general model queries handled through Graph-RAG, upward tracing of failure consequences, and downward identification of minimal success paths using the LLM agent tools. In this architecture, the language model interprets user queries and invokes predefined graph-based tools, while probabilistic and logical reasoning remains governed by the encoded model structure. This separation helps maintain traceability and limits unsupported reasoning during interaction.

Overall, the findings indicate that retrieval-based construction generates consistent DML representations from technical documentation while preserving explicit logical relationships. Although further validation across additional systems and functional equivalence testing would strengthen generalization, the framework provides a practical method for assisting DML development and evaluating structural quality in a systematic manner.

**CRediT Authorship Contribution Statement**

Saman Marandi: Methodology, Software, Formal analysis, Investigation, Data curation, Visualization, Writing - original draft.

Yu-Shu Hu: Conceptualization, Methodology, Writing - review & editing.

Mohammad Modarres: Conceptualization, Methodology, Resources, Supervision, Project administration, Writing - review & editing.

**Data Availability Statement**

The code supporting this study will be made publicly available upon acceptance of the manuscript. Additional data and materials are available from the authors upon request.

**Declaration of Competing Interest**

The authors declare that they have no known competing financial interests or personal relationships that could have appeared to influence the work reported in this paper.

**Funding**

This research did not receive any specific grant from funding agencies in the public, commercial, or non-profit sectors.

**Declaration of AI Use**

During the preparation of this work, the authors used ChatGPT 5.3 and Claude Sonnet 4.6 to assist with improving clarity and grammar of the manuscript. The authors reviewed and edited all content as needed and take full responsibility for the content of the published article.

## References

[1] I. A. Papazoglou, “Mathematical foundations of event trees,” *Reliability Engineering & System Safety*, vol. 61, no. 3, pp. 169–183, Sep. 1998, doi: 10.1016/S0951-8320(98)00010-6.

[2] M. E. Paté-Cornell, “Fault Trees vs. Event Trees in Reliability Analysis,” *Risk Analysis*, vol. 4, no. 3, pp. 177–186, Sep. 1984, doi: 10.1111/j.1539-6924.1984.tb00137.x.

[3] S. Kabir, “An overview of fault tree analysis and its application in model based dependability analysis,” *Expert Systems with Applications*, vol. 77, pp. 114–135, Jul. 2017, doi: 10.1016/j.eswa.2017.01.058.

[4] E. Ruijters and M. Stoelinga, “Fault tree analysis: A survey of the state-of-the-art in modeling, analysis and tools,” *Computer Science Review*, vol. 15–16, pp. 29–62, Feb. 2015, doi: 10.1016/j.cosrev.2015.03.001.

[5] M. Lind, *Foundations for Functional Modeling of Technical Artefacts*, Cham, Switzerland: Springer, 2024, doi: 10.1007/978-3-031-45918-4.

[6] M. Modarres, “Functional modeling of complex systems using a GTST-MPLD framework,” in *Proc. Int. Workshop on Functional Modeling of Complex Technical Systems*, Ispra, Italy, May 1993.

[7] Y.-S. Hu and M. Modarres, “Evaluating system behavior through Dynamic Master Logic Diagram (DMLD) modeling,” *Reliability Engineering & System Safety*, vol. 64, no. 2, pp. 241–269, May 1999, doi: 10.1016/S0951-8320(98)00066-0.

[8] Y.-S. Hu and M. Modarres, “Time-dependent system knowledge representation based on dynamic master logic diagrams,” *Control Engineering Practice*, vol. 4, no. 1, pp. 89–98, Jan. 1996, doi: 10.1016/0967-0661(95)00211-5.

[9] Y.-S. Hu and M. Modarres, “Logic-Based Hierarchies for Modeling Behavior of Complex Dynamic Systems with Applications,” in *Fuzzy Systems and Soft Computing in Nuclear Engineering*, vol. 38, D. Ruan, Ed., Heidelberg, Germany: Physica-Verlag, 2000, pp. 364–395, doi: 10.1007/978-3-7908-1866-6_17.

[10] Y.-S. Hu and M. Modarres, “Apply Fuzzy-Logic-Based Functional-Center Hierarchies as Inference Engines for Self-Learning Manufacture Process Diagnoses,” in *Fuzzy Systems and Knowledge Discovery*, vol. 3614, L. Wang and Y. Jin, Eds., Berlin, Germany: Springer, 2005, pp. 1012–1021, doi: 10.1007/11540007_129.

[11] W. X. Zhao *et al.*, “A Survey of Large Language Models,” *arXiv preprint arXiv:2303.18223*, Mar. 2025, doi: 10.48550/arXiv.2303.18223.

[12] C. Cuskley, R. Woods, and M. Flaherty, “The Limitations of Large Language Models for Understanding Human Language and Cognition,” *Open Mind*, vol. 8, pp. 1058–1083, Aug. 2024, doi: 10.1162/opmi_a_00160.

[13] M. Hutson, “Words vs. worlds,” Science, vol. 392, no. 6805, pp. 1336–1339, Jun. 2026, doi: 10.1126/science.aej9814.

[14] S. Marandi, Y.-S. Hu, and M. Modarres, “Complex System Diagnostics Using a Knowledge Graph-Informed and Large Language Model-Enhanced Framework,” *Applied Sciences*, vol. 15, no. 17, p. 9428, Aug. 2025, doi: 10.3390/app15179428.

[15] S. Ji, S. Pan, E. Cambria, P. Marttinen, and P. S. Yu, “A Survey on Knowledge Graphs: Representation, Acquisition and Applications,” *IEEE Transactions on Neural Networks and Learning Systems*, vol. 33, no. 2, pp. 494–514, Feb. 2022, doi: 10.1109/TNNLS.2021.3070843.

[16] H.-L. Zhu *et al.*, “A new risk assessment method based on belief rule base and fault tree analysis,” *Proceedings of the Institution of Mechanical Engineers, Part O: Journal of Risk and Reliability*, vol. 236, no. 3, pp. 420–438, Jun. 2022, doi: 10.1177/1748006X211011457.

[17] P. Weber and L. Jouffe, “Complex system reliability modelling with Dynamic Object Oriented Bayesian Networks (DOOBN),” *Reliability Engineering & System Safety*, vol. 91, no. 2, pp. 149–162, Feb. 2006, doi: 10.1016/j.ress.2005.03.006.

[18] M. Modarres and S. W. Cheon, “Function-centered modeling of engineering systems using the goal tree–success tree technique and functional primitives,” *Reliability Engineering & System Safety*, vol. 64, no. 2, pp. 181–200, May 1999, doi: 10.1016/S0951-8320(98)00062-3.

[19] J. Wu, X. Zhang, M. Song, and M. Lind, “Challenges in Functional Modelling for Safety and Risk Analysis,” in *Proceedings of the 33rd European Safety and Reliability Conference (ESREL 2023)*, M. P. Brito, T. Aven, P. Baraldi, M. Čepin, and E. Zio, Eds., Singapore: Research Publishing, 2023, pp. 1892–1899, doi: 10.3850/978-981-18-8071-1_P132-cd.

[20] M. Modarres, “Functional modeling of complex systems with applications,” in *Proc. Annual Reliability and Maintainability Symposium (RAMS)*, Washington, DC, USA: IEEE, 1999, pp. 418–425, doi: 10.1109/RAMS.1999.744153.

[21] Z. Hao, F. Di Maio, and E. Zio, “A sequential decision problem formulation and deep reinforcement learning solution of the optimization of O&M of cyber-physical energy systems (CPESs) for reliable and safe power production and supply,” *Reliability Engineering & System Safety*, vol. 235, p. 109231, Jul. 2023, doi: 10.1016/j.ress.2023.109231.

[22] D. T. Chung, M. Modarres, and R. N. M. Hunt, “GOTRES: An expert system for fault detection and analysis,” *Reliability Engineering & System Safety*, vol. 24, no. 2, pp. 113–137, Jan. 1989, doi: 10.1016/0951-8320(89)90088-4.

[23] Z. Hao, F. Di Maio, and E. Zio, “Dynamic Reliability Assessment of Cyber-Physical Energy Systems (CPESs) by GTST-MLD,” in *Proc. 5th Int. Conf. System Reliability and Safety (ICSRS)*, Palermo, Italy: IEEE, Nov. 2021, pp. 98–102, doi: 10.1109/ICSRS53853.2021.9660671.

[24] M. Modarres, D. Marksberry, T. Ballard, and V. Krivtsov, “Reactor safety assessment systems: Summary of methods and experience,” in *Proc. PSAM III / ESREL ’96 Conf.*, Crete, Greece, Jun. 1996.

[25] C. Guo, S. Gong, L. Tan, and B. Guo, “Extended GTST-MLD for aerospace system safety analysis,” *Risk Analysis*, vol. 32, no. 6, pp. 1060–1071, Jun. 2012, doi: 10.1111/j.1539-6924.2011.01718.x.

[26] R. Pennings, M. Ponamalé, and G. Gerlinger, “A methodology for the construction of safety-oriented advisory systems for operators,” *International Journal of Industrial Ergonomics*, vol. 17, no. 4, pp. 367–374, Apr. 1996, doi: 10.1016/0169-8141(95)00061-5.

[27] V. Garg, M. Prasad, G. Vinod, and J. Chattopadhyay, “Reliability analysis of smart pressure transmitter,” in *Reliability, Safety and Hazard Assessment for Risk-Based Technologies*, P. V. Varde, R. V. Prakash, and G. Vinod, Eds., Singapore: Springer, 2020, pp. 133–141, doi: 10.1007/978-981-13-9008-1_11.

[28] Y. F. Li, S. Valla, and E. Zio, “Reliability assessment of generic geared wind turbines by GTST-MLD model and Monte Carlo simulation,” *Renewable Energy*, vol. 83, pp. 222–233, Nov. 2015, doi: 10.1016/j.renene.2015.04.035.

[29] F. Brissaud, A. Barros, C. Berenguer, and D. Charpentier, “Reliability study of an intelligent transmitter,” in *Proc. 15th ISSAT Int. Conf. Reliability and Quality in Design*, Aug. 2009.

[30] M. Modarres and N. Kececi, “Software development life cycle model to ensure software quality,” in *Proc. PSAM IV Conf.*, New York, NY, USA, Sep. 1998.

[31] M. Modarres and Y.-S. Hu, “Fuzzy hierarchical modeling for business environment scanning and decision making,” in *Proc. PSAM 9 Conf.*, Hong Kong, China, May 2008.

[32] F. Antonello, J. Buongiorno, and E. Zio, “A methodology to perform dynamic risk assessment using system theory and modeling and simulation: Application to nuclear batteries,” *Reliability Engineering & System Safety*, vol. 228, p. 108769, Dec. 2022, doi: 10.1016/j.ress.2022.108769.

[33] R. H. Tai *et al.*, “An examination of the use of large language models to aid analysis of textual data,” *International Journal of Qualitative Methods*, vol. 23, p. 16094069241231168, Jan. 2024, doi: 10.1177/16094069241231168.

[34] J. Dagdelen *et al.*, “Structured information extraction from scientific text with large language models,” *Nature Communications*, vol. 15, no. 1, p. 1418, Feb. 2024, doi: 10.1038/s41467-024-45563-x.

[35] L. Huang *et al.*, “A survey on hallucination in large language models: Principles, taxonomy, challenges, and open questions,” *ACM Transactions on Information Systems*, vol. 43, no. 2, pp. 1–55, Mar. 2025, doi: 10.1145/3703155.

[36] S. Farquhar, J. Kossen, L. Kuhn, and Y. Gal, “Detecting hallucinations in large language models using semantic entropy,” *Nature*, vol. 630, no. 8017, pp. 625–630, Jun. 2024, doi: 10.1038/s41586-024-07421-0.

[37] Y. Gao *et al.*, “Retrieval-augmented generation for large language models: A survey,” *arXiv preprint arXiv:2312.10997*, 2023, doi: 10.48550/arXiv.2312.10997.
[38] Y. Li, X. Fu, G. Verma, P. Buitelaar, and M. Liu, “Mitigating hallucination in large language models (LLMs): An application-oriented survey on RAG, reasoning, and agentic systems,” *arXiv preprint arXiv:2510.24476*, 2025, doi: 10.48550/arXiv.2510.24476.
[39] J. Chen, H. Wu, J. Pang, Y. Wang, D. Zhang, and C. Sun, “Tool learning with language models: A comprehensive survey of methods, pipelines, and benchmarks,” *Vicinagearth*, vol. 2, no. 1, p. 16, Nov. 2025, doi: 10.1007/s44336-025-00024-x.
[40] J. Wang and V. X. Wang, “Assessing consistency and reproducibility in the outputs of large language models: Evidence across diverse finance and accounting tasks,” *SSRN preprint*, 2025, doi: 10.2139/ssrn.5189069.
[41] S. Pratap, A. R. Aranha, D. Kumar, G. Malhotra, A. P. N. Iyer, and S. S. S., “The fine art of fine-tuning: A structured review of advanced LLM fine-tuning techniques,” *Natural Language Processing Journal*, vol. 11, p. 100144, Jun. 2025, doi: 10.1016/j.nlp.2025.100144.
[42] B. Chen, Z. Zhang, N. Langrené, and S. Zhu, “Unleashing the potential of prompt engineering for large language models,” *Patterns*, vol. 6, no. 6, p. 101260, Jun. 2025, doi: 10.1016/j.patter.2025.101260.
[43] S. Zheng, K. Pan, J. Liu, and Y. Chen, “Empirical study on fine-tuning pre-trained large language models for fault diagnosis of complex systems,” *Reliability Engineering & System Safety*, vol. 252, p. 110382, Dec. 2024, doi: 10.1016/j.ress.2024.110382.
[44] J. Zhang, C. Zhang, J. Lu, and Y. Zhao, “Domain-specific large language models for fault diagnosis of heating, ventilation, and air conditioning systems by labeled-data-supervised fine-tuning,” *Applied Energy*, vol. 377, p. 124378, Jan. 2025, doi: 10.1016/j.apenergy.2024.124378.
[45] H. A. A. M. Qaid, B. Zhang, D. Li, S.-K. Ng, and W. Li, “FD-LLM: Large language model for fault diagnosis of machines,” *arXiv preprint arXiv:2412.01218*, 2024, doi: 10.48550/arXiv.2412.01218.
[46] J. Wang, T. Li, Y. Yang, S. Chen, and W. Zhai, “DiagLLM: Multimodal reasoning with large language model for explainable bearing fault diagnosis,” *Science China Information Sciences*, vol. 68, no. 6, p. 160103, Jun. 2025, doi: 10.1007/s11432-024-4333-7.
[47] Y. Lai, Z. Wu, M. Chen, C. Liu, and H. Shao, “FR-LLM: Multi-task large language model with signal-to-text encoding and adaptive optimization for joint fault diagnosis and RUL prediction,” *Reliability Engineering & System Safety*, vol. 269, p. 112091, May 2026, doi: 10.1016/j.ress.2025.112091.
[48] K. M. Alsaif, A. A. Albeshri, M. A. Khemakhem, and F. E. Eassa, “Multimodal large language model-based fault detection and diagnosis in the context of Industry 4.0,” *Electronics*, vol. 13, no. 24, p. 4912, Dec. 2024, doi: 10.3390/electronics13244912.
[49] X. Y. Lee, L. Vidyaratne, A. Farahat, and C. Gupta, “Exploring LLM-based agentic frameworks for fault diagnosis,” *Proc. Annual Conf. PHM Society*, vol. 17, no. 1, Oct. 2025, doi: 10.36001/phmconf.2025.v17i1.4350.
[50] Y. Gu *et al.*, “Argos: Agentic time-series anomaly detection with autonomous rule generation via large language models,” *arXiv preprint arXiv:2501.14170*, 2025, doi: 10.48550/arXiv.2501.14170.
[51] Y. Liu, Y. Zhou, Y. Liu, Z. Xu, and Y. He, “Intelligent fault diagnosis for CNC through the integration of large language models and domain knowledge graphs,” *Engineering*, vol. 53, pp. 311–322, Oct. 2025, doi: 10.1016/j.eng.2025.04.003.
[52] Y. Lan, M. Zhang, M. Su, and F. Zhou, “Knowledge-graph-enhanced and LLM-guided fault diagnosis for VSC-HVDC systems,” *AIP Advances*, vol. 15, no. 11, p. 115330, Nov. 2025, doi: 10.1063/5.0309229.
[53] X. Yang *et al.*, “CTI-Thinker: An LLM-driven system for CTI knowledge graph construction and attack reasoning,” *Cybersecurity*, vol. 9, no. 1, p. 106, Jan. 2026, doi: 10.1186/s42400-025-00505-y.
[54] C. Dong, D. Li, and H. R. Karimi, “Reinforcement learning driven adaptive graph construction for fault diagnosis of chemical processes,” *Reliability Engineering & System Safety*, vol. 266, p. 111781, Feb. 2026, doi: 10.1016/j.ress.2025.111781.

[55] J. Liang, H. Meng, and Y. Mu, "Domain-specific large language model-driven risk analysis of battery energy storage systems," *Reliability Engineering & System Safety*, vol. 274, p. 112416, Oct. 2026, doi: 10.1016/j.ress.2026.112416.
[56] H. Zhang, Y. Zhao, B. Sun, Y. Wu, Z. Fu, and X. Xiao, "Large Language Model Based Intelligent Fault Information Retrieval System for New Energy Vehicles," *Appl. Sci.*, vol. 15, no. 7, p. 4034, Apr. 2025, doi: 10.3390/app15074034.
[57] W. Lin and K. Miao, "An automotive fault diagnosis framework based on knowledge graphs and large language models," *Electronics*, vol. 14, no. 21, p. 4180, Oct. 2025, doi: 10.3390/electronics14214180.
[58] P. Liu, L. Qian, X. Zhao, and B. Tao, "Joint knowledge graph and large language model for fault diagnosis and its application in aviation assembly," *IEEE Transactions on Industrial Informatics*, vol. 20, no. 6, pp. 8160–8169, Jun. 2024, doi: 10.1109/TII.2024.3366977.
[59] J. Xu, Z. Chen, H. Ren, Z. Jiang, Y. Wang, and W. Gui, "Text-augmented contrastive evaluation method: Pioneeringly achieving quantitative assessment for RAG-enhanced LLM of industrial fault diagnosis," *SSRN preprint*, 2025, doi: 10.2139/ssrn.5146755.
[60] S. Marandi, Y.-S. Hu, and M. Modarres, "Integrating large language models and knowledge graphs for system diagnostics," in *Proc. Annual Reliability and Maintainability Symposium (RAMS)*, Miramar Beach, FL, USA: IEEE, Jan. 2026, pp. 1–6, doi: 10.1109/RAMS50514.2026.11424548.
[61] S. Marchesin, G. Silvello, and O. Alonso, "Large language models and data quality for knowledge graphs," *Information Processing & Management*, vol. 62, no. 6, p. 104281, Nov. 2025, doi: 10.1016/j.ipm.2025.104281.
[62] H. Huang, C. Chen, Z. Sheng, Y. Li, and W. Zhang, "Can LLMs be good graph judges for knowledge graph construction?," *arXiv preprint arXiv:2411.17388*, 2024, doi: 10.48550/arXiv.2411.17388.
[63] P. Plamper, H. Köpcke, and A. Groß, "A survey on spatio-temporal knowledge graph models," *arXiv preprint arXiv:2512.16487*, 2025, doi: 10.48550/arXiv.2512.16487.
[64] Neo4j, Inc., "Neo4j GitHub repository," *GitHub*. [Online]. Available: https://github.com/neo4j/neo4j. Accessed: Nov. 2025.
[65] K. Palacio-Rodríguez, I. Lans, C. N. Cavasotto, and P. Cossio, "Exponential consensus ranking improves the outcome in docking and receptor ensemble docking," *Scientific Reports*, vol. 9, no. 1, p. 5142, Mar. 2019, doi: 10.1038/s41598-019-41594-3.
[66] W. Wu and G. Konstantinidis, "Compliance as a trust metric," *arXiv preprint arXiv:2601.01287*, 2026, doi: 10.48550/arXiv.2601.01287.
[67] J. J. Curry, D. W. Gallagher, M. Modarres, and J. A. Radder, *Interim reliability-evaluation program: Analysis of the Millstone Point Unit 1 nuclear power plant. Volume I: Main report*, U.S. Nuclear Regulatory Commission, NUREG/CR-3085/1, May 1983.
[68] D. Kančev and M. Čepin, "A new method for explicit modelling of single failure event within different common cause failure groups," *Reliab. Eng. Syst. Saf.*, vol. 103, pp. 84–93, Jul. 2012, doi: 10.1016/j.ress.2012.03.009.
[69] V. Hassija, C. Senthil Kumar, and K. Velusamy, "A pragmatic approach to estimate alpha factors for common cause failure analysis," *Ann. Nucl. Energy*, vol. 63, pp. 317–325, Jan. 2014, doi: 10.1016/j.anucene.2013.07.053.
[70] W. E. Vesely, M. Belhadj, and J. T. Rezos, "PRA importance measures for maintenance prioritization applications," *Reliab. Eng. Syst. Saf.*, vol. 43, no. 3, pp. 307–318, Jan. 1994, doi: 10.1016/0951-8320(94)90035-3.

## Appendix A

This appendix provides full per-run evaluation results for all node-level and link-level metrics across five executions of the model. These results are included to ensure transparency and to support detailed inspection of where errors occur within the hierarchical structure. Link-level relationships are reported between adjacent layers of the hierarchy, including Goal-Function, Function-Subfunction, Subfunction-Component, and Component-Success Condition connections. Node-level results are reported for the Goals, Functions, Subfunctions, Components, and Success Conditions layers. The evaluation metrics reported include precision, recall, and the $F_2$ score for both node-level and link-level results. In addition, gate accuracy is reported for link-level relationships and measures whether the logical operator assigned to each parent group matches the reference model. All metrics are bounded between 0 and 1, where a value of 1 indicates perfect performance.

**Table A1**. Node-Level Precision.

| Run | Goals | Functions | Subfunctions | Components | Success Conditions |
|---|---|---|---|---|---|
| Run 1 | 1.000 | 1.000 | 1.000 | 1.000 | 1.000 |
| Run 2 | 1.000 | 1.000 | 1.000 | 1.000 | 1.000 |
| Run 3 | 1.000 | 1.000 | 1.000 | 0.992 | 1.000 |
| Run 4 | 1.000 | 1.000 | 1.000 | 0.983 | 1.000 |
| Run 5 | 1.000 | 1.000 | 1.000 | 1.000 | 1.000 |

**Table A2**. Node-Level Recall.

| Run | Goals | Functions | Subfunctions | Components | Success Conditions |
|---|---|---|---|---|---|
| Run 1 | 1.000 | 1.000 | 1.000 | 0.975 | 0.975 |
| Run 2 | 1.000 | 1.000 | 1.000 | 0.983 | 0.983 |
| Run 3 | 1.000 | 1.000 | 1.000 | 0.975 | 0.983 |
| Run 4 | 1.000 | 1.000 | 1.000 | 0.992 | 0.992 |
| Run 5 | 1.000 | 1.000 | 1.000 | 0.983 | 0.983 |

**Table A3.** Node-Level $F_2$ Score.

| Run | Goals | Functions | Subfunctions | Components | Success Conditions |
|---|---|---|---|---|---|
| Run 1 | 1.000 | 1.000 | 1.000 | 0.980 | 0.980 |
| Run 2 | 1.000 | 1.000 | 1.000 | 0.987 | 0.987 |
| Run 3 | 1.000 | 1.000 | 1.000 | 0.978 | 0.987 |
| Run 4 | 1.000 | 1.000 | 1.000 | 0.990 | 0.993 |
| Run 5 | 1.000 | 1.000 | 1.000 | 0.987 | 0.987 |

**Table A4.** Link-Level Precision.

| Run | Goal-Function | Function-Subfunction | Subfunction-Component | Component-Success Condition |
|---|---|---|---|---|
| Run 1 | 1.000 | 1.000 | 0.994 | 1.000 |
| Run 2 | 1.000 | 1.000 | 1.000 | 1.000 |
| Run 3 | 1.000 | 1.000 | 1.000 | 1.000 |
| Run 4 | 1.000 | 1.000 | 0.989 | 1.000 |
| Run 5 | 1.000 | 1.000 | 0.995 | 1.000 |

**Table A5.** Link-Level Recall.

| Run | Goal-Function | Function-Subfunction | Subfunction-Component | Component-Success Condition |
|---|---|---|---|---|
| Run 1 | 1.000 | 1.000 | 0.962 | 0.975 |
| Run 2 | 1.000 | 1.000 | 0.951 | 0.983 |
| Run 3 | 1.000 | 1.000 | 0.962 | 0.983 |
| Run 4 | 1.000 | 1.000 | 0.973 | 0.992 |
| Run 5 | 1.000 | 1.000 | 0.978 | 0.983 |

**Table A6.** Link-Level $F_2$ Score.

| Run | Goal-Function | Function-Subfunction | Subfunction-Component | Component-Success Condition |
|---|---|---|---|---|
| Run 1 | 1.000 | 1.000 | 0.968 | 0.980 |
| Run 2 | 1.000 | 1.000 | 0.961 | 0.987 |
| Run 3 | 1.000 | 1.000 | 0.969 | 0.987 |
| Run 4 | 1.000 | 1.000 | 0.976 | 0.993 |
| Run 5 | 1.000 | 1.000 | 0.982 | 0.987 |

**Table A7.** Link-Level Gate Accuracy

| Run | Goal-Function | Function-Subfunction | Subfunction-Component | Component-Success Condition |
|---|---|---|---|---|
| Run 1 | 1.000 | 1.000 | 0.926 | 1.000 |
| Run 2 | 1.000 | 0.889 | 0.962 | 1.000 |
| Run 3 | 1.000 | 1.000 | 0.963 | 1.000 |
| Run 4 | 1.000 | 1.000 | 0.889 | 1.000 |
| Run 5 | 1.000 | 1.000 | 0.926 | 1.000 |